%% file: main.tex
\documentclass{article} 
\usepackage{arxiv, times} 
\usepackage{natbib}

\usepackage{hyperref}

\usepackage{url}
\usepackage{todonotes}

\usepackage{xargs}

\usepackage{marginnote}

\usepackage[font=small,labelfont=bf]{caption}

\usepackage{xargs}

\usepackage{marginnote}

\newcommand{\bx}{\mathbf{x}}
\newcommand{\by}{\mathbf{y}}
\newcommand{\bX}{\mathbf{X}}

\input{math_commands.tex}

\usepackage{hyperref}
\hypersetup{
    colorlinks=true,
    linkcolor=red,
    filecolor=magenta,
    urlcolor=blue,
    citecolor=purple,
    pdftitle={Overleaf Example},
    pdfpagemode=FullScreen,
    }

\usepackage[capitalize,noabbrev]{cleveref}

\title{Disentangling the Causes of Plasticity Loss in Neural Networks}

\author{Clare Lyle\thanks{Correspondence to \texttt{clarelyle@google.com}}, Zeyu Zheng, Khimya Khetarpal, Will Dabney, Hado van Hasselt, Razvan Pascanu, James Martens \\
Google DeepMind\\
}

\begin{document}

\maketitle

\begin{abstract}
    Underpinning the past decades of work on the design, initialization, and optimization of neural networks is a seemingly innocuous assumption: that the network is trained on a \textit{stationary} data distribution. In settings where this assumption is violated, e.g.\ deep reinforcement learning, learning algorithms become unstable and brittle with respect to hyperparameters and even random seeds. 
    One factor driving this instability is the loss of plasticity, meaning that updating the network's predictions in response to new information becomes more difficult as training progresses. While many recent works provide analyses and partial solutions to this phenomenon, a fundamental question remains unanswered: to what extent do known mechanisms of plasticity loss overlap, and how can mitigation strategies be combined to best maintain the trainability of a network?
    This paper addresses these questions, showing that loss of plasticity can be decomposed into multiple independent mechanisms and that, while intervening on any single mechanism is insufficient to avoid the loss of plasticity in all cases, intervening on multiple mechanisms in conjunction results in highly robust learning algorithms. We show that a combination of layer normalization and weight decay is highly effective at maintaining plasticity in a variety of synthetic nonstationary learning tasks, and further demonstrate its effectiveness on naturally arising nonstationarities, including reinforcement learning in the Arcade Learning Environment.
\end{abstract}

\section{Introduction}

Training a neural network on a single task is relatively straightforward: for popular data modalities it is usually possible to take an off-the-shelf architecture and optimization algorithm and, with minimal hyperparameter tuning, obtain a reasonably accurate model. Yet the statistical relationships learned by machine learning systems are often dynamic: consumer preferences change over time, information on the internet goes out of date, and words enter and leave common usage. In reinforcement learning systems, the very act of improving the learner's policy 
introduces changes in the distribution of data it collects for further training. Without the ability to update its predictions in response to these changes, a learning system's performance will inevitably decline.
Such declines in performance are often resolved in practice by resetting the model parameters and training from scratch on updated training data, but this solution introduces additional computational overhead which can be prohibitively expensive for large models. It is therefore desirable to avoid the need for parameter resets by maintaining the network's ability to adapt to new learning signals throughout the course of training.

The observation that training a neural network on some learning problems results in a reduced ability to adapt to new tasks has been made independently several times, both in reinforcement learning \citep{dohare2021continual, igl2021transient, kumar2020implicit, lyle2021understanding} and in supervised learning \citep{ash2020warm, berariu2021study}. 
While this phenomenon, which we will refer to in this paper as \textit{plasticity loss}, is quite prevalent, our understanding of why it occurs remains vague.
\citet{lyle2023understanding} show a number of negative results indicating that we cannot attribute all instances of plasticity loss to a single measured quantity such as parameter norm or number of dead units. As a result, methods which aim to preserve plasticity typically either target a single pathology \citep{sokar2023dormant, abbas2023loss}, which runs the risk of allowing plasticity loss to occur via other unaddressed mechanisms, or regularize more abstract properties of the network towards their initial value \citep{lyle2021understanding, lewandowski2023curvature, kumar2023maintaining}, which can run the risk of interfering with the primary objective. 
A useful model of plasticity loss should combine the best of both of these approaches, allowing for the development of principled mitigation strategies which target several independent mechanisms in conjunction. 

This paper aims to develop such a model, and use it to guide the development of more effective methods to maintain plasticity.
We begin with an empirical analysis which explores three questions: first, what types of nonstationarity induce plasticity loss? Second, what types of structural changes occur to the parameters and features of neural networks when they lose plasticity? Third, what properties are shared among networks which have lost plasticity? Our analysis reveals surprising answers to these questions, finding that multiple distinct mechanisms of plasticity loss can be unified by a underlying phenomenon of preactivation distribution shift, and result in similar degeneracies in the network's empirical neural tangent kernel (which are predictive of training pathologies). While some of the mechanisms we identify are already known, such as dead units, we identify some surprising new ones, such as unit linearization. 
We furthermore identify the importance of the magnitude of the targets in regression tasks, finding that target magnitude alone can explain a significant fraction of previous examples of plasticity loss in deep reinforcement learning.  While we do not aim to provide an exhaustive list of all network pathologies that can lead to loss of trainability, we show that plasticity loss can in fact be driven by several \textit{independent} mechanisms, in the sense that an intervention which acts on one such mechanism may still allow plasticity loss to occur by another.

We proceed to take these insights to develop a `Swiss cheese model' of mitigation strategies, and show that interventions which improve robustness to preactivation distribution shift, regression target magnitude, and parameter growth can be studied independently and then combined to give additive benefit. Whereas prior work indicated that each mechanism on its own cannot fully explain plasticity loss \citep{lyle2023understanding}, we find that addressing several mechanisms in conjunction is sufficient to yield negligible loss of plasticity in a variety of synthetic benchmarks, and that these methods can also improve performance on both deep reinforcement learning benchmarks and on natural distribution shift datasets. By identifying the best interventions for each mechanism in isolation and then combining them, we can significantly reduce the combinatorial complexity of the search over mitigation strategies, a finding which has exciting implications for future work on the stabilization of optimization in nonstationary learning problems.
\section{Background and related work}
Our analysis in this paper is grounded in two related branches of the neural network literature. The first is the study of what properties make a network trainable at initialization. The second is a study of how networks can deviate from these trainable initializations over the course of training, and as a result lose plasticity. 

\subsection{Network training, signal propagation, and preactivation distributions}
\label{sec:sig-prop}

The training dynamics of neural networks have been extensively studied over the past decades \citep[e.g.][]{sutskever2013importance}. This literature has highlighted a number of important design choices for neural network architectures, initialization schemes, and optimizers, which will come to bear on the analysis in this paper. Of particular relevance to this work is layer normalization \citep{ba2016layer}, which normalizes vectors in a network (often preactivations) to have sample mean 0 and variance 1, commonly used activation functions like ReLU and tanh, and schemes designed achieve good signal propagation\footnote{Roughly speaking, networks with poor signal propagation do not do a good job of propagating information about their inputs to deeper layers. This can lead to optimization difficulties, such as vanishing or exploding gradients, as has been formalized in \citet{xiao2020disentangling} and \citet{martens2021rapid} in the ``NTK regime".} via careful initialization \citep{poole2016exponential,balduzzi2017shattered} or network design and transformation \citep{martens2021rapid,hayou2021stable,zhang2021deep,he2023deep}.

In the signal propagation literature \citep[e.g][]{daniely2016towards,poole2016exponential,schoenholz2017deep,lee2017deep,xiao2018dynamical,hanin2019finite, hayou2019impact,yang2019wide,martens2021rapid}, the distribution of preactivations (i.e.\ the inputs to element-wise nonlinear layers) plays a crucial role in predicting the functional behavior, and ultimate trainability, of a randomly initialized network. In the most popular schemes to achieve good signal propagation, parameters are initialized as i.i.d. Gaussian with mean and variance chosen so that the empirical preactivation distribution within a layer resembles a Gaussian with some target mean $\mu$ and variance $\sigma^2$ (with high probability for all network inputs). The choices of $\mu$ and $\sigma^2$ are crucial, and their optimal values will depend on the characteristics of the activation function, as well as those of the network's architecture \citep{martens2021rapid}. 
While there can be no universally optimal choice\footnote{To see this, note that if $\mu_0$ and $\sigma_0^2$ are optimal for a given activation $\phi_0$, then $\mu_1 = \mu_0 + 3$ and $\sigma_1^2 \equiv \sigma_0^2 / 10$ would be optimal for the activation defined by $\phi_1(x) \equiv \phi_0(\sqrt{10} \, (x - 3))$.} for $\mu$ and $\sigma^2$, the standard choice $\mu = 0$ and $\sigma^2 = 1$ will often be good enough for many commonly used activation functions, assuming modest depth or the use of residual connections \citep{he2016deep}.
\citet{martens2021rapid} and \citet{zhang2021deep} even give methods for transforming any activation function to work well with $\mu = 0$ and $\sigma^2 = 1$.
Notably, applying layer normalization to a preactivation vector ensures that the 1st and 2nd order sample statistics precisely agree with $\mu = 0$ and $\sigma^2 = 1$. While this would not guarantee the maintenance of good signal propagation as training progresses, as this depends on various other conditions holding, it may indeed be helpful. Indeed, prior works \citep{lyle2023understanding, kumar2023continual} have observed that networks which incorporate normalization layers tend to be better able to maintain plasticity.

\subsection{Loss of plasticity}
A good initialization ensures that, at least when training begins, the network it is applied to is capable of learning -- in other words, that it is \textit{plastic}. As training progresses, however, it is possible for this plasticity to be lost \citep{dohare2021continual, lyle2021understanding, berariu2021study, abbas2023loss}. The term plasticity has been used to refer to both a network's ability to attain low generalization error~\citep{berariu2021study}, and its ability to improve its performance on a training set~\citep{abbas2023loss, nikishin2023deep, kumar2023maintaining}. This paper will adopt the latter interpretation, focusing on mitigating the optimization difficulties observed in nonstationary learning problems. 
Specifically, in this work we will say that a network has lost plasticity if it is unable to optimize its objective function as effectively as a randomly initialized network. Plasticity can thus be thought of as the quality of a particular point in parameter space to serve as a starting point for optimization. It can be useful to also include the optimizer state in this definition, as illustrated by \citet{asadi2023resetting}. In this paper, we use the terms plasticity and trainability interchangeably, in line with the usage of \citet{lyle2023understanding} as opposed to \citet{berariu2021study}. Notably, this definition does not imply that a loss of plasticity is permanent -- it is certainly in principle possible for the network to enter an ill-conditioned region of the loss landscape where optimization is slow and then escape. Even a transient reduction in trainability, however, can have serious implications on the performance of learning algorithms in contexts such as deep RL, and so it is desirable to avoid these regions.

Several works have studied the mechanisms of plasticity loss and proposed novel mitigation strategies \citep{ash2020warm, dohare2021continual, nikishin2022primacy, abbas2023loss, lee2023plastic}. In reinforcement learning, additional analysis has gone into identifying instabilities due to bootstrapping which may exacerbate this phenomenon \citep{kumar2020implicit}. While some prior works have suggested that TD targets may induce overfitting in neural networks \citep{raileanu2021decoupling, lyle2022learning}, growth of their scale over the course of training has not been considered as a possible impediment to optimization outside of multi-task settings \citep{van2016learning}. Despite several analyses proposing potential explanations of plasticity loss, the work of \citet{lyle2023understanding} suggests that none of these mechanisms presents a causal explanation of plasticity loss in isolation.
\section{A deeper look into plasticity}

\label{sec:two-pathologies}
In this section we aim to decompose the process of plasticity loss into three phases: the learning dynamics induced by a task, the effect of these dynamics on the parameters and features of the network, and the manifestation of these changes in networks which have lost plasticity. We will conclude from this analysis that while plasticity loss can occur via a variety of independent mechanisms, it tends to produce similar pathologies in the empirical neural tangent kernel (which, as we will argue, are predictive of training difficulties).

\begin{figure}
    \centering
    \begin{minipage}{0.538\textwidth}
    \centering
    \includegraphics[width=\linewidth]{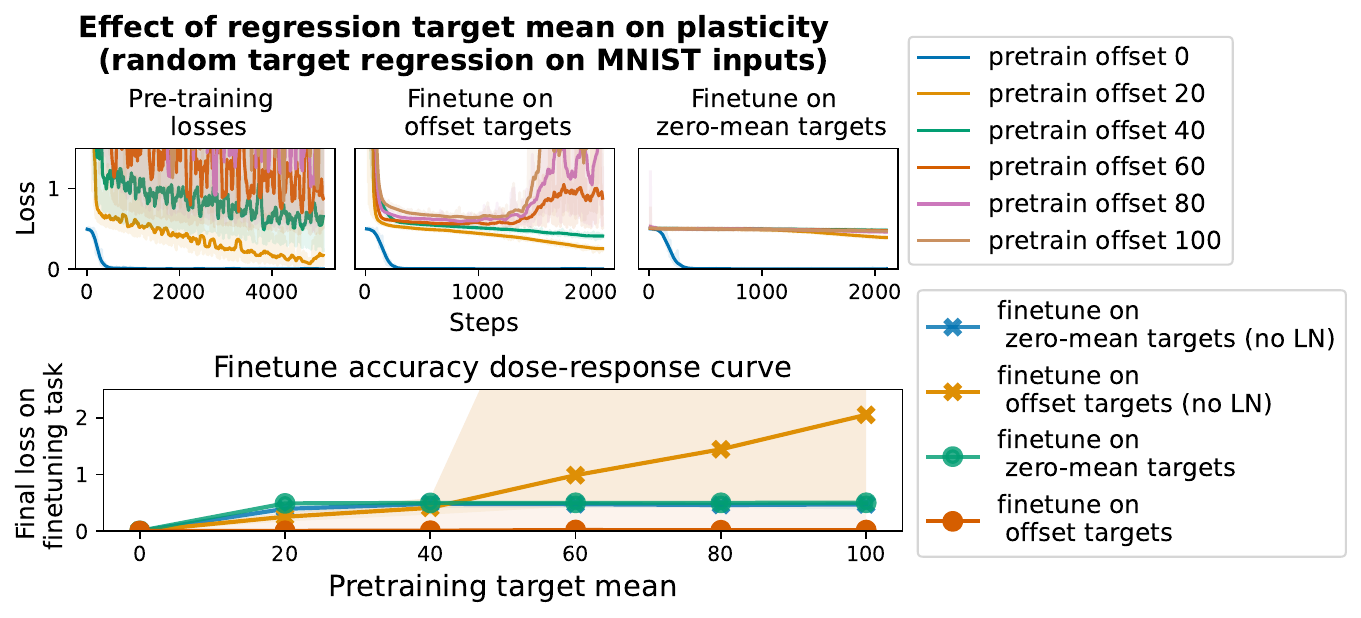}
    
    \end{minipage}
    \begin{minipage}{0.448\textwidth}
    \centering
    \includegraphics[width=\linewidth]{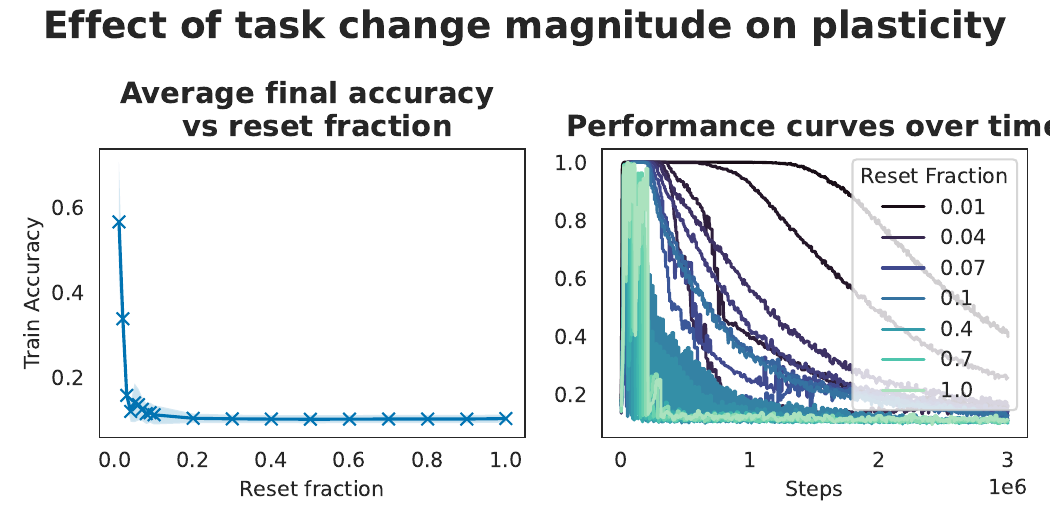}
    
    \end{minipage}
    \caption{ \textbf{Left.} Illustration of the relationship between pretraining target magnitude and optimization speed on a new task. We see a strong dose-response effect from increasing the magnitude of regression targets on the final loss on a fine-tuning task, and observe similar trends in the learning curves on these tasks as those observed by \citet{lyle2023understanding} in DQN agents trained on contextual bandits. \textbf{Right.} Dose-response curves for the effect of target offset scale and the magnitude of the distribution shift on plasticity. We see that more severe distribution shifts, e.g. randomizing an entire dataset simultaneously rather than gradually, results in more extreme loss of plasticity.} \vspace{-1em}
    \label{fig:param-growth}
\end{figure}

\subsection{What kinds of learning problems induce plasticity loss?}
There are many situations in which neural networks do not lose plasticity in a way which interferes with the minimization of their objective function -- if this were not the case, we would not have seen the widespread adoption of deep learning in fields such as computer vision and natural language processing. This then raises the question: what properties of a learning problem \textit{do} cause a neural network to lose plasticity?
In this subsection, we outline two independent factors which can be demonstrated to induce plasticity loss in neural networks: the magnitude of the regression target mean in regression problems, and the smoothness of the distribution shift induced by the nonstationarity. Though we do not claim that these two factors exclusively characterize all problems which might impair trainability, their independence is an important theme that will recur throughout this paper.

\textbf{Regression target scale.} 
The first property we will consider is motivated by the analysis of \citet{lyle2023understanding}, who observed increased optimization difficulty as a function of training steps in a DQN agent trained on a trivially simple contextual bandit environment, whose observation space was an image classification dataset, and whose reward function gave reward 1 whenever the selected action corresponded to the label of the observation. The resulting nonstationarity induced by training a DQN agent on such a task is highly structured and essentially amounts to regressing on one-hot image labels and periodically adding a constant offset to the regression targets. We claim that it is this deceptively simple structure to the nonstationarity which drove the optimization difficulties encountered by this agent, i.e. that the nonstationarity \textit{increases the target mean}, rather than the existence of nonstationarity in the first place. To prove this, we show that similar loss of plasticity can be induced by simply regressing neural networks on targets with large means in a \textit{stationary} learning problem.

To do so, we construct a fixed regression problem consisting of random output targets offset by some fixed constant, using as input images from the MNIST dataset. We use this learning problem as a ``pretraining" task, and then ``fine-tune" the network on a new set of random target outputs, offset by either the same constant or zero. We observe in Figure~\ref{fig:param-growth} (leftmost column and top right subplot) that pretraining on large offsets significantly reduces the network's ability to learn on new targets, whether these new targets have the same mean as the pretraining task or are centered at zero, and that the degree of interference with new tasks increases as a function of the pretraining offset. The size of the offset used in the pretraining task is inversely correlated with the network's ability to fit later tasks, exhibiting a dose-response effect. While layer normalization can mitigate this trend, it does not completely eliminate it, particularly in the case of centered fine-tuning targets. In Appendix~\ref{sec:scale-plasticity} we provide additional analysis which suggests that the primary mechanism for this loss of plasticity is that the network learns to encode the offset in a single-dimensional subspace of the feature space, leading to poorly-conditioned feature matrices as the bias encoding dwarfs the other components of the features. As we will see in Section~\ref{sec:loss-characterization}, an analogous phenomenon occurs in the empirical NTK of networks trained on these tasks as well.

\textbf{Smoothness of distribution shifts.} A simple nonstationary task in which one can quickly induce loss of plasticity is the sequential memorization of randomized labels. It has been argued in prior work that the sudden distribution shift in gradient magnitudes induced by these rapid task changes is responsible for much of the catastrophic loss of plasticity which has been observed in these domains~\citep{lyle2023understanding}. We construct a simple experiment to explore how sensitive the plasticity loss phenomenon is to the suddenness of this distribution shift by training a CNN (architecture details in Appendix~\ref{appx:experiment-details}) on a sequence of randomly generated labels of the CIFAR-10 image classification dataset. At each iteration, we re-randomize some fraction $\epsilon \in [0.01, 1]$ of these labels, and continue training from the network's current parameters. We train each network for the equivalent of 40 iterations of 100,000 steps (applying resets more frequently for smaller re-randomization fractions to attain the same total `amount' of label resets). We use three activation functions, ReLU, GeLU, and Leaky ReLU (with negative slope equal to 0.01).

We see in the top leftmost subplot of Figure~\ref{fig:param-growth} that more sudden task changes result in more severe loss of plasticity, as evidenced by the downward trajectory as the fraction of the dataset which is reset grows in the top leftmost inset. In the right hand side, we visualize the trajectories of these networks over several iterations, observing that larger re-randomization fractions lead to more precipitous declines in networks. Notably, resetting only 1\% of labels minimally interferes with performance. Increasing this fraction to 10\% rapidly accelerates the decline, with a slightly smoother decrease but otherwise parallel trend in the network with leaky ReLU activations.
In the next section, we will explore the implications of these observations and take a fine-grained look into precisely how label re-randomization causes networks to become less trainable.

\subsection{Mechanisms of plasticity loss}
\label{sec:mechanisms}

In some cases the mechanism by which a network has lost plasticity is obvious, for example when all hidden units are saturated. In other cases, the mechanisms by which the network becomes less trainable are harder to pin down. In this subsection, we identify two independent mechanisms by which plasticity can be lost. The first of these mechanisms concerns distribution shifts in the preactivations, which can result in the known pathology of unit death, but also more subtle problems we identify, such as degraded signal propagation and \textit{linearization} of units. The second concerns the norm of the parameters of the model, which has downstream effects on the sharpness of the loss landscape and the sensitivity of the network output to changes in the weights.

\begin{figure}
    \centering
    \includegraphics[height=5.0cm]{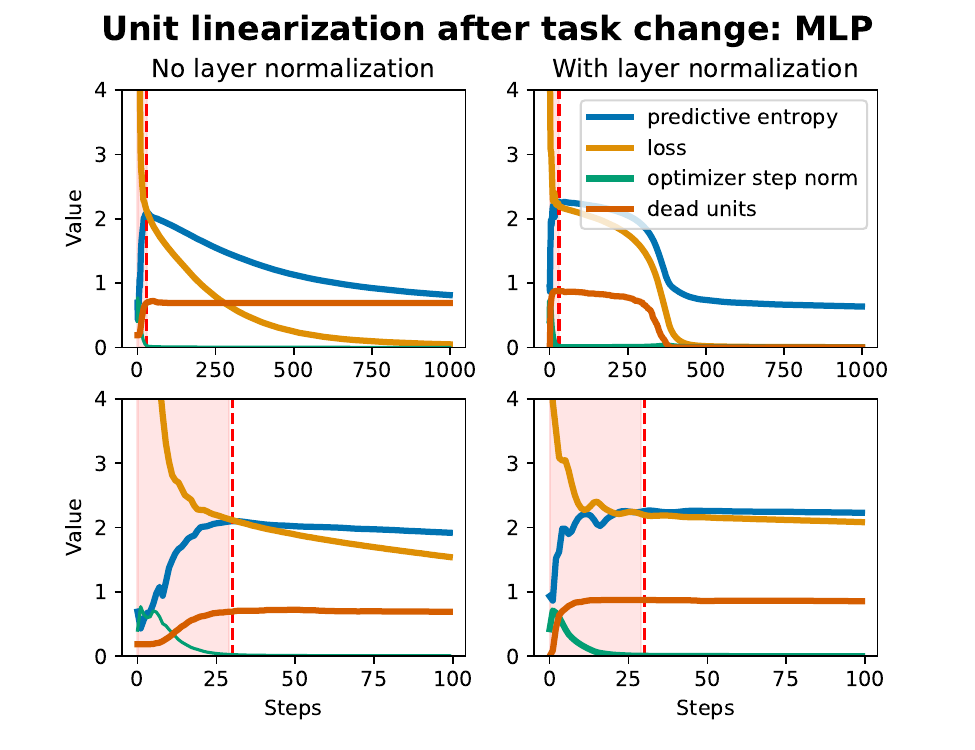}
    \includegraphics[height=5.0cm]{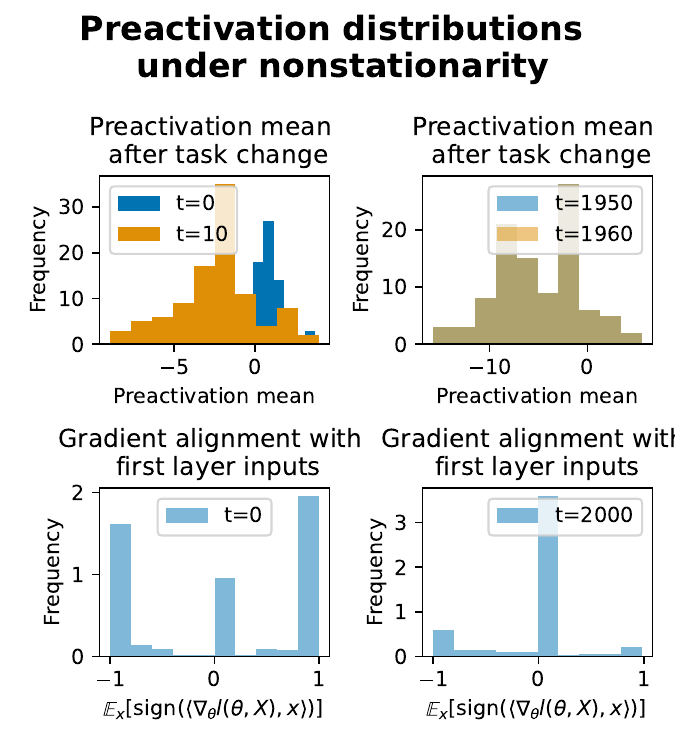}
    \includegraphics[height=5.0cm]{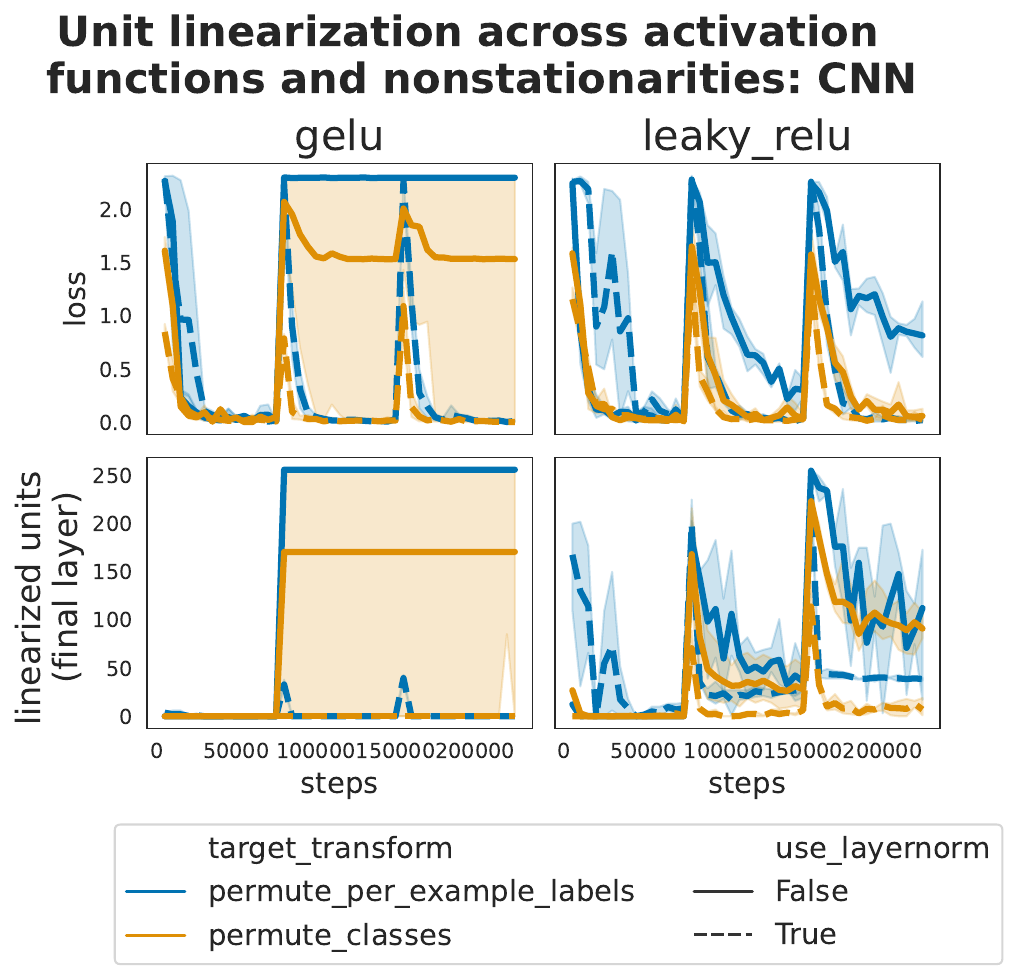}
    \caption{\textbf{Accumulation of dead units after a task change:} we visualize the learning dynamics of a small MLP trained to memorize random labels of the MNIST dataset immediately after a task change. \textbf{Left.} We highlight an early phase of training wherein predictive entropy sharply increases along with the number of dead units. \textbf{Middle.} The distribution shift in the pre-activations in this early phase, and the concomitant spike in gradients with negative dot product with \textit{all} incoming features at the start of learning, contrasting with the stable dynamics later in training. \textbf{Right.} Similar spikes in linearized units occur immediately after a task change in convolutional networks trained with different activation functions on the CIFAR=10 dataset.} \vspace{-0.5em}
    \label{fig:dead-and-zombies}
\end{figure}

\textbf{Linearization and preactivation distribution shift.}
\label{sec:dead-units}
As discussed previously, a network consisting primarily of `dead' \citep{lin2016far} or `dormant' \citep{sokar2023dormant} units will not be trainable.
A dead unit is unable to propagate gradients back to its incoming weights, which means that unless the unit's input distribution shifts in a suitable direction, the parameters associated with this unit will remain permanently frozen. (For example, a ReLU might have all negative preactivation values and thus always output 0, or a tanh might have very large preactivation values that take it into its ``constant" region.)
One simple solution to dead units is to use a non-saturating activation function, such as Leaky ReLU. However, changes to the mean or variance (across the training set) of the preactivation for a unit can still lead to pathologies. For example, assuming the initial preactivation distribution is chosen to achieve good signal propagation (as discussed in Section \ref{sec:sig-prop}), large changes to it may harm signal propagation, leading to negative downstream effects on optimization dynamics.

Beyond harming signal propagation, a preactivation distribution change may result in more easily measurable pathologies. One such pathology is the \textit{linearization} of a unit, whereby the unit acts as a linear (or nearly linear) transformation of its inputs, effectively reducing the expressivity of the network. For example, a ReLU unit with only positive inputs will behave like the identity, or any smooth activation unit with very low variance inputs will behave close to a linear function. Unlike dead units, through which gradients cannot flow by definition, these `zombie units' can propagate gradients (and indeed permit ``perfect" signal propagation); however, the presence of too many of them will reduce the \textit{effective} expressive power of the network \citep[e.g.][]{martens2013representational,montufar2014, raghu2017expressive}.
While saturated units have received much attention as a factor in plasticity loss, the accumulation of effectively-linear units has not previously been studied in the context of network plasticity.

To study how training dynamics might drive such pre-activation distribution shifts immediately after a sudden change to the training objective, we train a neural network to convergence on one set of random labels of MNIST, and then re-shuffle the labels and continue optimization from the converged parameters and optimizer state. We observe in Figure~\ref{fig:dead-and-zombies} two phases of training dynamics after a task change: first an `erasing' phase, where confidence in incorrect predictions is rapidly reduced and predictive entropy increases, followed by a `disentanglement' phase where the network adapts to fit the structure of the new task and predictive entropy declines. These two phases are exemplified with a marked increase in the number of dead units immediately after the task switch, followed by a re-equilibration period during which some units desaturate. We illustrate these two phases in the left-hand-side plot in Figure \ref{fig:dead-and-zombies}, where we track the loss and number of dead units in a small MLP immediately after a task change in the random labels task. 
We see that initially the updates on the incoming weights to the first layer tend to either increase or decrease the preactivation values for \textit{all training inputs to the unit}. This can be explained by the pressure from the loss function in the erasing phase to reduce the magnitude of incorrect logits and increase the entropy of the predicted distribution (see Figure~\ref{fig:unit-death-norm} in Appendix D for further details). Coupled with large gradient magnitudes, and hence large step sizes immediately after a task switch, this results in a high rate of units entering the linearized regime, either saturated or unsaturated. As the optimization dynamics shift towards increasing the logits of correct labels, the gradient directions become more diverse, and the nonlinear structure of the network recovers at least partially, provided a sufficient fraction of the hidden units are not saturated.

\begin{figure}
    \centering
    \includegraphics[height=3.85cm]{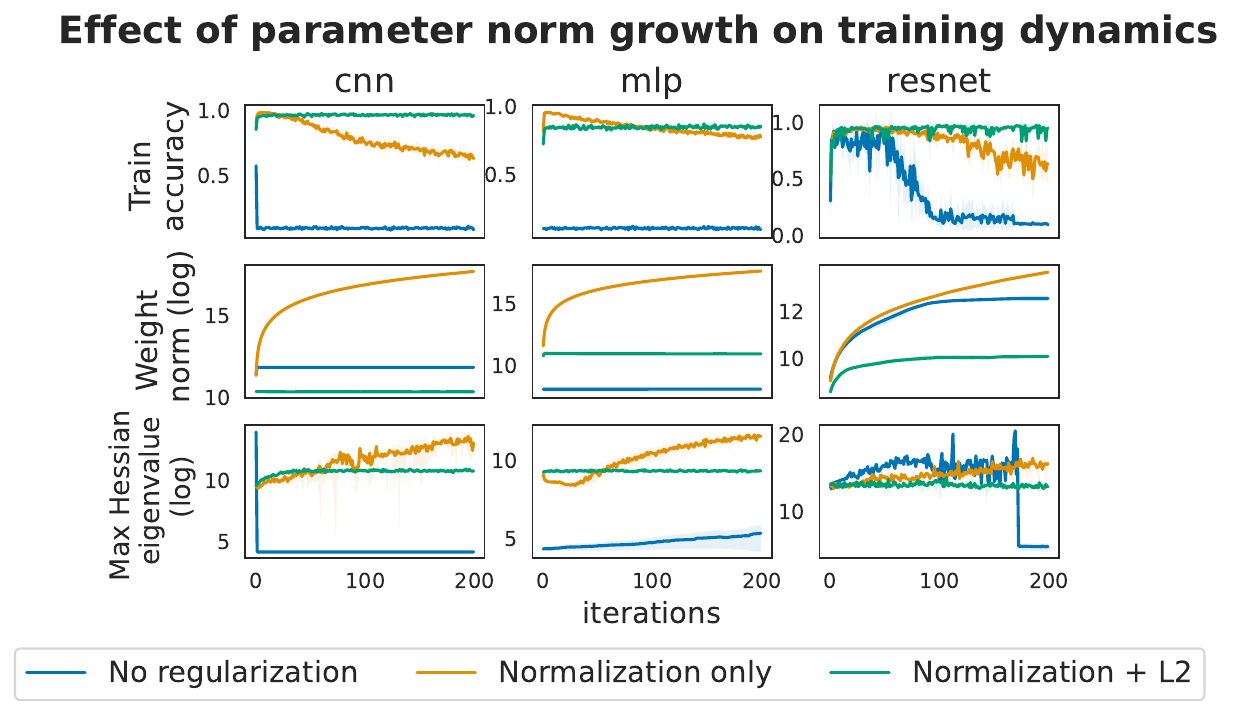}
    \includegraphics[height=3.8cm]{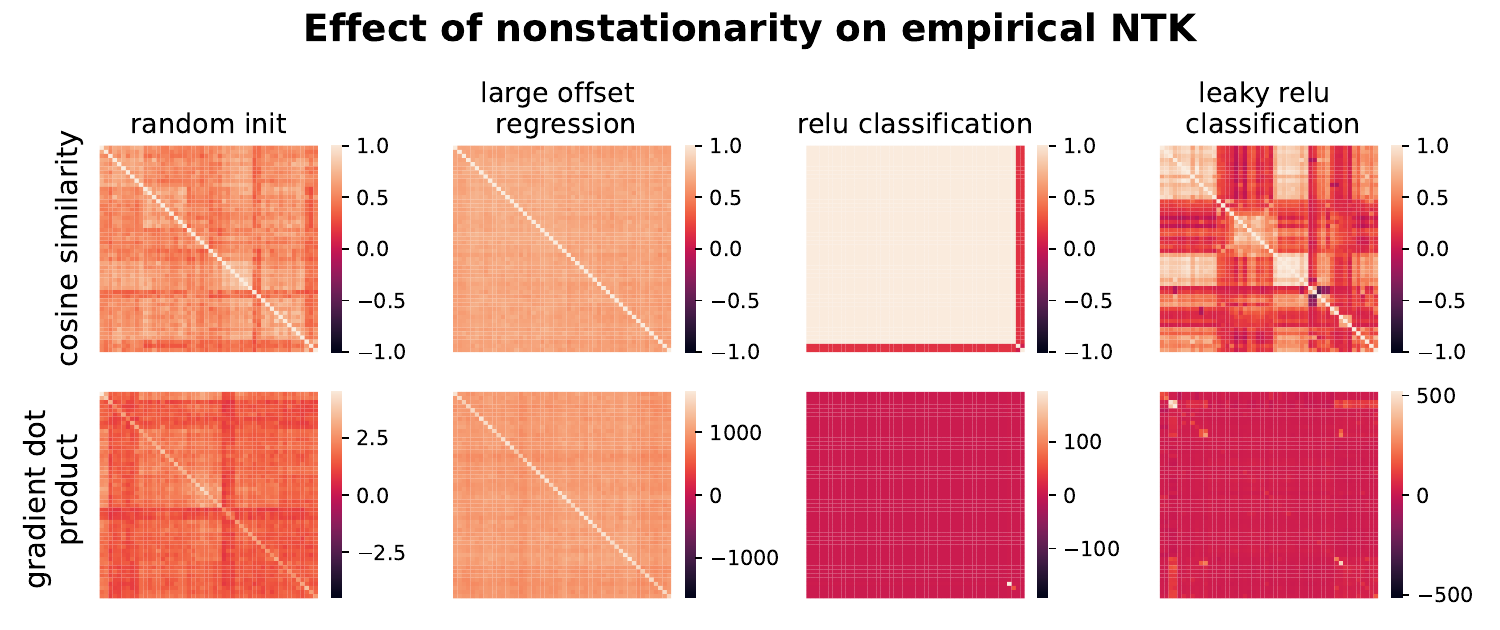}
    \caption{\textbf{Left.} Visualization of the effect of parameter scale on learning. \textbf{Right.} Visualization of the empirical NTKs of a variety of networks which exhibit varying degrees of plasticity loss, taken from previously-studied scenarios. We observe greater interference in networks which have lost plasticity than in a random initialization (i.e. greater magnitude of the cosine similarities in the top row), and also exhibit greater gradient norm variance across inputs.} \vspace{-0.5em}
    \vspace{-0.5em}
    \label{fig:ntk-scale-figure}
\end{figure}

\textbf{Parameter norm growth.} 
We observe two main learning difficulties associated with growth in the network parameters. The first is training instabilities, which can take the form of numerical overflow errors if parameter norm is allowed to grow indefinitely, but also more subtle pathologies. As we see in Figure~\ref{fig:ntk-scale-figure}, parameter norm tends to correlate with the sharpness as measured by the Hessian. Such a connection between loss landscape sharpness and last-layer parameter norm is explored in the context of edge-of-stability dynamics by \citet{damian2022selfstabilization}. A more subtle issue can also arise when the norms of different layers grow at different rates. Many recent works on the infinite-width limit propose scaling the learning rates for different network layers in proportion to their incoming and outgoing dimensions as a means of ensuring that updates to each layer have an equal effect on the network output \citep{jacot2018neural, yang2019wide}. However, as we demonstrate in Appendix~\ref{sec:unequal-param-growth}, it is not uncommon for parameters to grow unevenly through the network, leading to learning difficulties as updates to each layer have differential effects on the network output.
The second concern is that increasing parameter magnitude can have the effect of saturating network components such as softmax attention heads \citep{wortsman2023small, dehghani2023scaling} or normalization layers \citep{merrill2021effects}, resulting in smaller changes in the network output for a fixed update step size than could be obtained at initialization.

We illustrate these two mechanisms of action in the iterated random label memorization tasks studied in the previous subsection by training a variety of architectures on iteratively re-randomized labels of CIFAR-10 for 15 million optimizer steps.
We see in the leftmost plot of Figure~\ref{fig:ntk-scale-figure} that increased parameter norm accompanies plasticity loss in all architectures we evaluate. In networks using normalization layers, the performance on new tasks degrades smoothly as the parameter norm grows. In networks which do not incorporate normalization layers, we see rapid early increases followed by plateaus due to saturated nonlinearities. Notably, there is not a monotone relationship between parameter norm at the end of training and plasticity: networks with all saturated units do not propagate gradients which could contribute to norm growth. This explains how parameter norm can be a \textit{causal} factor in plasticity loss, while also not exhibiting a monotonic relationship. 

\subsection{Characterizing networks which have lost plasticity}
\label{sec:loss-characterization}

Thus far, we have identified a variety of external and internal factors influencing plasticity in a neural network. We now explore whether these different paths ultimately take networks to the same end point. To do so, we look at the structure of the gradients of the network's outputs (not its loss) with respect to its parameters using the empirical neural tangent kernel (NTK) of the network. The empirical NTK matrix (eNTK for short) is the matrix of gradient dot products between all data points in the training set, a matrix which is both expensive to compute and difficult to visualize in its full resolution.  In practice, we sample a mini-batch of size 64 and compute the matrix of gradient dot products $\langle \nabla_\theta f(\bx), \nabla_\theta f(\bx') \rangle$ for all inputs $\bx, \bx'$ in the mini-batch. 

We visualize these matrices as heat maps in Figure~\ref{fig:ntk-scale-figure}. To interpret the heat maps, it is helpful to think of the empirical NTK as decomposing into the sum of two matrices: a diagonal matrix $D_\theta$ and a non-diagonal matrix $G_\theta$. Intuitively, the structure and magnitude of $G_\theta$ characterizes how the network generalizes. A network which does not generalize at all between inputs would have a diagonal eNTK, i.e. $G_\theta = 0$. In contrast, a network which multiplies the input by zero and then adds a learnable constant can be expressed as $G_\theta = c \mathbf{1}$ for some $c$ where $\mathbf{1}$ is the all-ones matrix, and $D_\theta = 0$, and its eNTK will have rank 1. In practice, it is desirable for the matrix $G_\theta$ to encode some geometric information about the inputs, so that the network generalizes more strongly between similar inputs and weakly between distant ones, although this is not critical for trainability~\citet{xiao2020disentangling}. 

The eNTK characterizes the local optimization dynamics \citep{jacot2018neural} of the network, analogous to the use of the infinite-width limit NTK to identify trainable initializations \citep{martens2021rapid}. An ill-conditioned eNTK serves as a signal that the network is suffering from optimization difficulties in the region of parameter space it has arrived at via its training trajectory.
We give a heuristic argument for why the eNTK might be of particular interest in the case of \textit{nonstationary} learning problems as follows. First, we recall that to a first-order approximation, the change in the loss obtained by following gradient descent on a dataset $\bX, \by$ with parameters $\theta$, function approximator $f_\theta: \Theta \times \mathbb{R}^d \rightarrow \mathbb{R}$, and objective $\ell(\theta) = \|f(\theta, \bX) - \by\|^2)$ can be expressed as
\begin{align*}
    \ell(\theta_t) - \ell(\theta_{t+1}) &\approx (f(\theta_t, \bX) - \by)^\top K_{\theta_t}(\bX, \bX) (f(\theta_t, \bX) - \by) + \mathrm{h.o.t.} 
\end{align*}
where $K_{\theta_t}(\bX, \bX)$ denotes the eNTK at timestep $t$. Supposing that $\by_t$ is dependent on $t$, then we cannot guarantee that a value of $K_{\theta_t}$ that was previously aligned with the residual $\by - f(\theta_t, \bX)$ will continue to have this property as $\by_t$ evolves. As a result, maintaining a non-collapsed NTK is of critical importance. 

We now revisit the conditions observed previously to induce training difficulties, and explore how loss of plasticity manifests in pathologies in the eNTK. We refer to Figure~\ref{fig:ntk-scale-figure}, which features two rows, the top of which visualizes the matrix of cosine similarities between gradients (i.e. it is the eNTK pre- and post-multiplied by a diagonal normalization matrix) and the bottom of which visualizes the eNTK itself. The leftmost column of this figure is a reference value for the empirical NTK of a randomly initialized network. In the second column is a network trained on a regression problem whose targets have a large mean relative to their variance. The eNTK of this network is close in $\ell_2$ distance to the sum of a diagonal matrix and a rank-1 matrix, despite having only trained for a few hundred steps. The third column visualizes the eNTK of a ReLU network trained to memorize a sequence of random labels. The network has accumulated a number of dead units, resulting in a block-diagonal structure. Replacing the ReLU activation with a Leaky ReLU in the rightmost column results in a similar, although less extreme, pathology. Strikingly, despite arising from completely different learning dynamics, all three of the networks which have loss plasticity tend to exhibit similar pathologies in their eNTK, centering around the proximity of the off-diagonal entries to a low rank matrix. While it is not straightforward to explicitly regularize the eNTK of a network during training, this observation points to the utility of the eNTK as a diagnostic tool for the loss of plasticity in neural networks.

\section{Mitigation strategies}
\label{sec:mitigation}
\begin{figure}
    \centering
    \includegraphics[width=\linewidth]{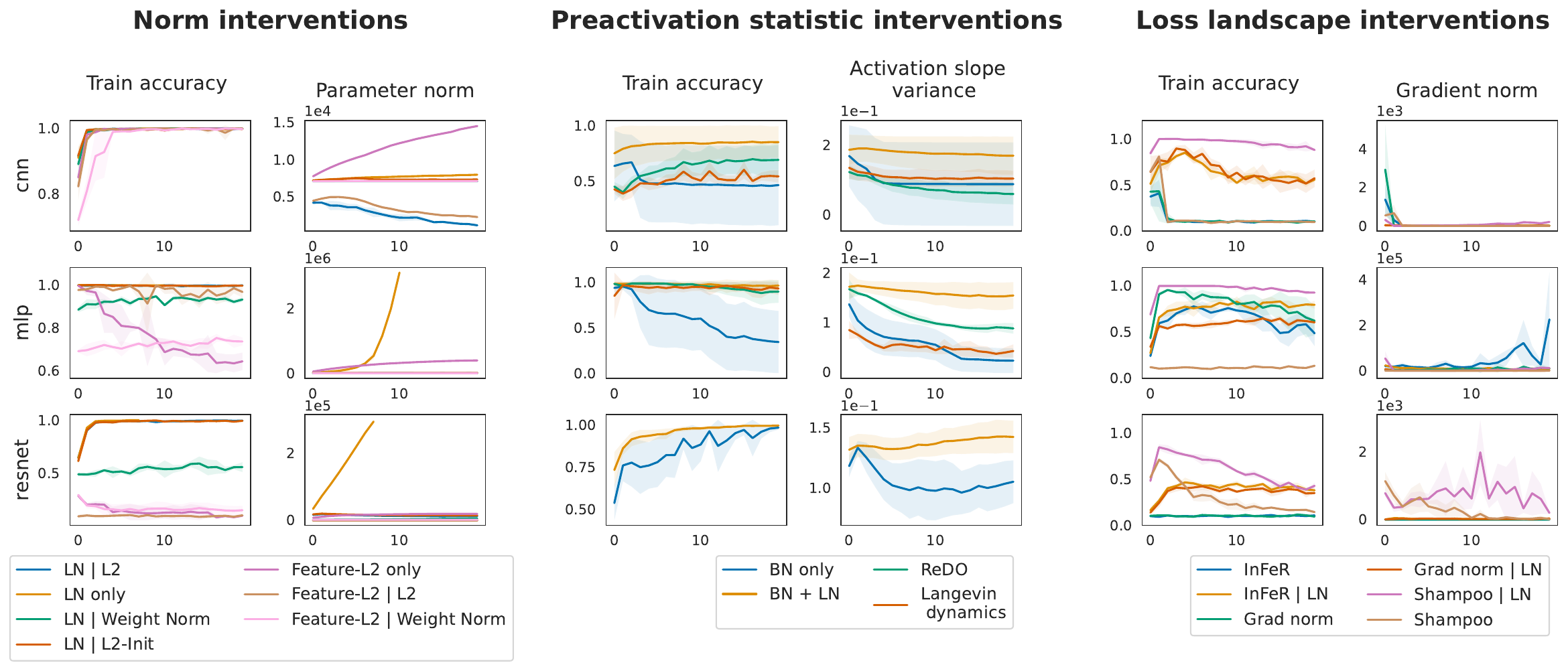}\vspace{-0.5em}
    \caption{Comparison of interventions aimed at addressing different failure modes: overall, combining layer normalization with L2 regularization addressed plasticity loss in all classification problems we considered. Many other strategies also improve over doing nothing, but do not outperform this baseline.} \vspace{-1em}
    \label{fig:intervention-sweep}
\end{figure}
The previous section identified a handful of mechanisms by which neural networks can lose plasticity: roughly speaking, these centered around growth of the parameter norm, distribution shift in the preactivations (leading to saturated, dead, and zombie units, among other possible pathologies associated with poor signal propagation), and poor conditioning of the features due to training with large target means.
In this section, we will investigate a variety of means by which these mechanisms can be intervened on to prevent plasticity loss. We then explore whether it is possible to combine the interventions for these independent mechanisms to achieve an additive benefit. For full details of the experiment settings in this section, we refer to Appendix~\ref{appx:experiment-details}.

\subsection{Addressing individual mechanisms}
\label{sec:param-feature-blowup}
\textbf{Unbounded norm growth.} To prevent feature and parameter norms from growing indefinitely, one can either enforce hard normalization constraints or use softer regularization strategies. We evaluate a variety of these in the leftmost subplot in Figure~\ref{fig:intervention-sweep}. We find that normalizing features along either the layer or batch dimension does not adversely affect network expressivity or learning speed; however, constraining the norm of the weights (in our case, by rescaling the per-layer vectorized weight norm to its value at initialization, a heuristic proxy to path normalization \citep{neyshabur2016path}) did interfere with optimization speed. Conversely, regularizing the norm of the features was less effective at combating plasticity loss than applying the hard constraint of layer or batch normalization.

\textbf{Preactivation statistics.}
One natural solution to the problem of dead units is to simply not use saturating activation functions \citep{abbas2023loss}; however, this strategy does not resolve the underlying problem of preactivation distribution shift. We consider two primary classes of interventions: constraining the pre-activations to have zero mean and unit variance along either the batch or layer dimension (using batch normalization and layer normalization respectively), and resetting units which have saturated using ReDO \citep{sokar2023dormant}.
We observe in the central subplot of Figure~\ref{fig:intervention-sweep} that normalizing the preactivations has beneficial effects on both plasticity in later task changes and on convergence rates in each task. In contrast, methods which reset dead or colinear features are more robust to plasticity loss than standard training approaches, but sometimes slow down convergence on single tasks and are unable to resolve signal propagation issues in larger networks that arise at initialization, which is why they are not included in our evaluations on ResNets. We include a more detailed sweep over more creative normalization layer combinations in Figure~\ref{fig:appx-normalization} of Appendix~\ref{sec:rebuttal}, finding slight improvements over the standard options considered here.

\textbf{Loss landscape conditioning.}
The regularization and normalization strategies studied previously occurred on the level of maintaining the expressivity of individual units. However, it is plausible that this unit-level intervention misses out on global properties of the loss landscape which may also benefit from regularization. We test this hypothesis in the rightmost subplot in Figure~\ref{fig:intervention-sweep}, where we consider a squared gradient norm penalty \citep{barrett2020implicit,smith2021origin}, regularization of a feature subspace towards its initial value \citep[InFeR,][]{lyle2021understanding}, and the curvature-aware optimizer Shampoo \citep{gupta2018shampoo,anil2020scalable}.
All results are run in networks with layer normalization but without L2 regularization. While we see some performance improvements at the end of training on some nonstationary tasks, none of these approaches consistently outperforms the combination of L2 regularization and layer normalization (of the pre-activations) studied earlier.

\begin{figure}
    \centering
    \includegraphics[width=0.63\linewidth]{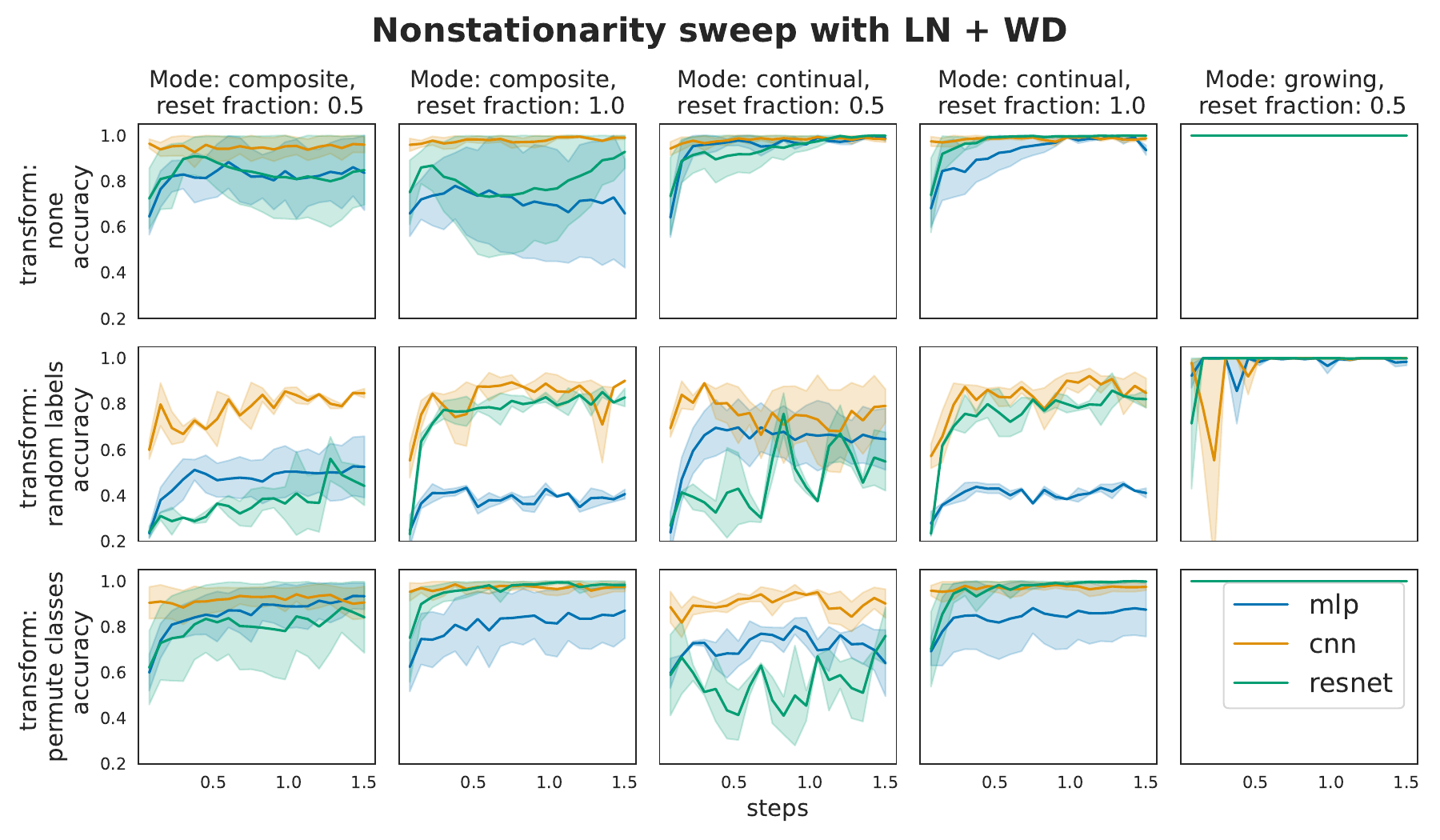}\vline
    \includegraphics[width=0.36\linewidth]{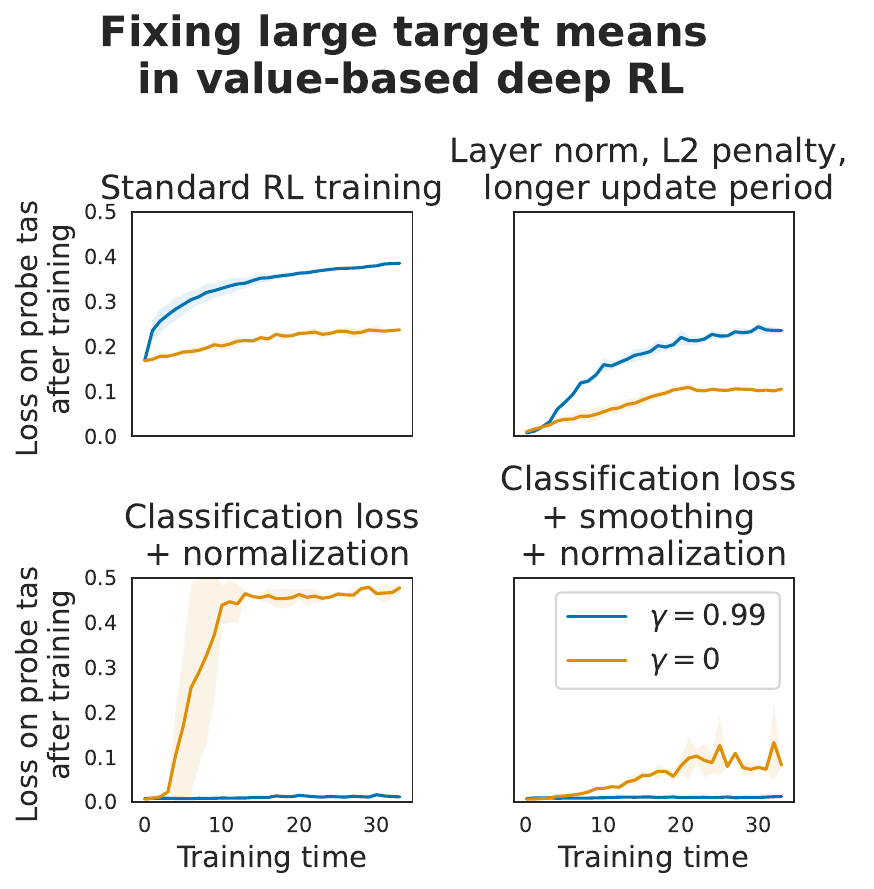}\vspace{-1em}
    \caption{\textbf{Left.} Layer normalization and L2 regularization on synthetic non-stationary supervised classification problems. \textbf{Right.} necessity of scale-invariant output parameterization in a simple RL task.}\vspace{-1em}
    \label{fig:method-synthetic}
\end{figure}

\textbf{Target scale.} While most of our previous evaluations focused on the classification regime, recall that one of the primary mechanisms by which we observed loss of plasticity was growth in the mean of the target distribution. We now take a small digression to revisit a reinforcement learning problem where plasticity loss was previously identified \citep{lyle2023understanding}. In this environment, the agent is randomly shown an observation from an image classification dataset and a reward of 1 is given if the action taken by the agent corresponds to the label of the image it observes. The resulting optimal value prediction problem amounts to predicting a one-hot vector corresponding to the class label of an image plus a constant bias term (whose value converges to roughly $(1-\gamma)^{-1}$). 
We train a DQN agent (see Appendix~\ref{appx:contextual-bandit} for details) on this task, and evaluate the plasticity of parameter checkpoints by regressing from those parameters to a fixed-norm perturbation of the network's current outputs. As before, the large target norms induced by the high discount factor result in plasticity loss even in the network incorporating layer normalization.

Distributional losses \citep{bellemare2017distributional,imani2018improving, schrittwieser2020mastering, stewart2023regression} have been demonstrated to improve performance on a variety of regression problems, and have been found to help mitigate plasticity loss \citep{lyle2023understanding}. In particular, we look at the `two-hot' trick (described in full detail in Appendix~\ref{appx:contextual-bandit}), which encodes the regression problem in terms of a categorical cross-entropy loss function, by placing a grid of points over the range of possible output values, and for each real-valued target, putting probability mass on the two closest points (proportional to its proximity to them). We see in the bottom left subplot of the right side of Figure \ref{fig:method-synthetic} that this appears to completely mitigate the problem for the $\gamma=0.99$ agent. The failure of $\gamma=0$ in the case of classification losses illustrates a more subtle issue which stems from the saturated softmax outputs, and consequent loss spike obtained when we evaluate the network on a new task, which we previously discussed in Section~\ref{sec:mechanisms}. This non-smoothness can be mediated by label smoothing (bottom right subplot) where the target distribution (defining the cross-entropy loss) is a mixture of a uniform distribution and the original two-hot encoding of the real-valued regression targets. This trick has the effect of scaling the mean of the target distribution by a constant factor, but because it does so uniformly for all inputs it does not interfere with accurate ranking and is easy to correct for if desired. Our findings here suggest an intriguing mechanism for some of the benefits of categorical distributional reinforcement learning: not only does the distributional loss provide smoother gradients \citep{imani2018improving}, but it also allows the network to represent large output values without developing ill-conditioned feature representations.

\subsection{Evaluation}
\label{sec:unified-optimizer}
The previous subsection observed that the combination of weight decay and layer normalization was highly effective at mitigating plasticity loss. This is in line with our previous discussion of the mechanisms that these two interventions influence: layer normalization benefits the preactivation distribution, while weight decay avoids catastrophic growth of the weight norm. We now explore how this combination fares under a broader range of nonstationarities.

\textbf{Supervised learning.} Consistently throughout our preceding evaluations, we have observed significant benefits from the simple combination of layer normalization and L2 regularization. We now expand our evaluations to incorporate input distribution shifts, gradually expanding datasets, and composite forms of nonstationarity where only a subset of inputs change over time. In each of the ``modes", re-randomization of the labels happens at a regular interval of 75k iterations, partitioning the training into a sequence of ``tasks".
We evaluate three architectures: a MLP, a CNN, and a ResNet-18 as detailed previously. Each architecture has layer normalization applied before each nonlinearity, and is trained with a fixed L2 penalty of $10^{-5}$ (additional tuning of the L2 penalty can yield better per-task performance, as can be seen in Figure~\ref{fig:l2_extra_sweep}). We use the CIFAR-10 benchmark as our base dataset. Each network is trained for 1.5M optimizer steps, which corresponds to 20 tasks. In the `continual' mode, a random transformation is applied to a given fraction of the dataset (chosen uniformly at random), either 0.5 or 1.0. In the `composite' mode, the random transformation is always applied to the same data points. In the `growing' mode, we start with inputs from a single class, and add inputs from an additional class at each task boundary, also applying random transformations to the data incorporated so far. 
Encouragingly, as the network trains on the sequence of 20 tasks we tend to see improving accuracy over time (left side of Figure \ref{fig:method-synthetic}), suggesting that provided mild guard rails are put in place, a number of network pathologies can be avoided while the network learns a useful inductive bias for the underlying task distribution. 

\textbf{Reinforcement learning.} 
Improving performance in RL requires striking a delicate balance between maintaining plasticity and not interfering with other aspects of the algorithm. Some regularization methods, such as L2 penalties and batch normalization, are known to interfere with learning in deep RL \citep{salimans2016weight}, constraining the set of reasonable interventions to consider. We therefore focus on applying layer normalization to the preactivations of a C51 and a Rainbow agent trained on the arcade learning environment. These agents combine both scale-invariant output encoding and the normalization approach discussed in Section~\ref{sec:mitigation}. 
\begin{figure}
    \centering
    \includegraphics[height=2.6cm]{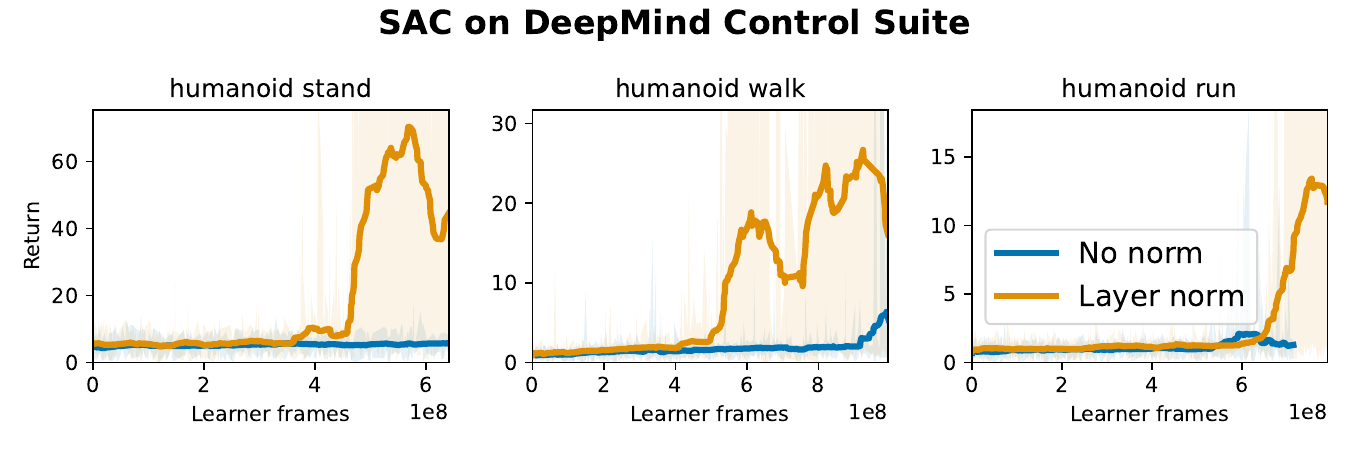} \vline
    \includegraphics[height=2.5cm]{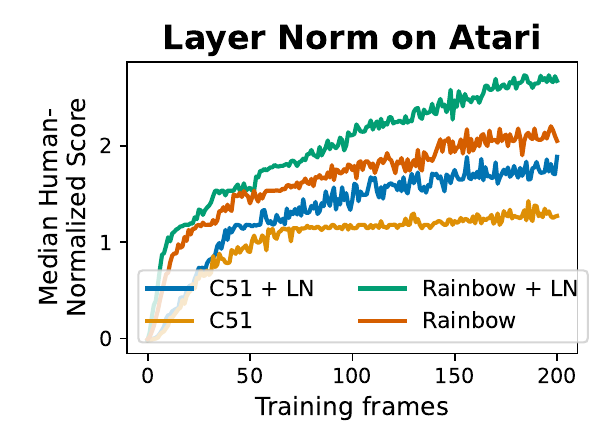} \vline \includegraphics[height=2.5cm]{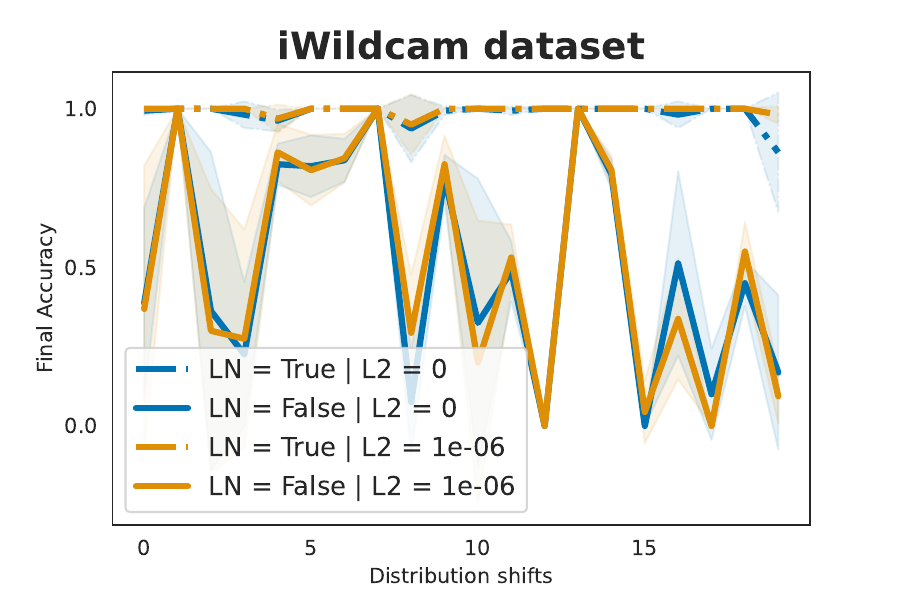} 
    \vspace{-0.5em}
    \caption{Left: performance of a soft actor-critic agent on the humanoid domain from the DeepMind Control Suite using architectures which include or exclude normalization layers. Middle: effect of incorporating layer normalization on value-based RL agents trained on all 57 games of the Arcade Learning Environment benchmark. Right: effect of normalization and L2 regularization on robustness to changes in photo location on the iWildcam dataset.}\vspace{-1em}
    \label{fig:natural-nonstats}
\end{figure}
We see in Figure~\ref{fig:natural-nonstats} that this approach does consistently provide modest performance improvements over equivalent baselines. We found that batch normalization and L2 regularization interfere with learning; these experiments and per-game performance of all agents can be found in Appendix~\ref{appx:detailed-atari}.
We also validate the beneficial effects of layer normalization on a Soft Actor-Critic agent trained on the DeepMind Control suite. We use fully connected architectures, applying layer normalization prior to each nonlinearity as in the Atari networks, with full details described in Appendix~\ref{appx:nonstat-benchmark-details}. Here again we find that layer normalization provides an almost uniform improvement over the suite. We see particularly striking improvements in the challenging \texttt{humanoid} domain as seen in Figure~\ref{fig:natural-nonstats}. Results for the full suite can be found in Appendix~\ref{sec:dm-control}.

\textbf{Natural distribution shifts.}
We further verify the robustness of our approach to natural distribution shifts using a dataset from the WiLDS benchmark \citep{koh2021wilds} which consists of nature camera photos taken from different locations and from different times of day. We follow an iterative training procedure, where we sample a location and train on images taken in that location for 2000 optimizer steps, then sample a new location and continue training on images from this location. We repeat this process 20 times, averaging over 5 random seeds.
Similarly to the synthetic tasks, we see in Figure~\ref{fig:natural-nonstats} that networks trained with L2 regularization and preactivation normalization are better able to deal with changes in the training data distribution, obtaining lower loss on later training sets than networks trained without these interventions. Although performance in the unregularized networks does not monotonically decline due to variable difficulty of different locations, we observe a widening gap between the final loss attained by the networks which use layer normalization and L2 regularization, confirming the efficacy of these interventions. A more detailed set of analyses can be found in Figure~\ref{fig:iwildcam-baseline}.

\vspace{-0.5em}

\section{Conclusions}
This paper has shown that while no single network property can explain all instances of plasticity loss in neural networks, a handful of independent mechanisms are responsible for a large fraction of observed cases. Some of these mechanisms are well-known to the community, such as dead ReLU units, but others such as large target offsets and preactivation distribution shift (resulting in unit ``zombification", among other problems) had not been previously identified and present exciting directions for future study, both as mechanisms of plasticity loss, and as potential targets for its remedies. We have further shown that by identifying effective mitigation strategies for each mechanism in isolation and then combining the most effective interventions, we can significantly reduce the combinatorial complexity of the search space of multi-component interventions. We anticipate that this divide-and-conquer strategy may prove useful to future works which aim to find even better mitigation strategies than those identified here. In particular, we note that while L2 regularization is effective at avoiding extreme weight norms (which can interfere with learning), it does so at the expense of convergence speed, and it's possible that better norm control strategies exist which do not interfere as much with single-task training speed.

\subsubsection*{Acknowledgments}
We thank Vincent Roulet, Mark Rowland, Diana Borsa, Alexandre Galashov, and many other colleagues at Google DeepMind for valuable discussions and feedback on this work.

\bibliography{main}
\bibliographystyle{plainnat}
\clearpage
\appendix
\input{appendix.tex}

\end{document}

%% file: math_commands.tex
\usepackage{amsmath,amsfonts,bm}

\def\eqref#1{equation~\ref{#1}}

\def\1{\bm{1}}

\DeclareMathAlphabet{\mathsfit}{\encodingdefault}{\sfdefault}{m}{sl}
\SetMathAlphabet{\mathsfit}{bold}{\encodingdefault}{\sfdefault}{bx}{n}



%% file: appendix.tex
\section{Additional background}
\subsection{Neural network training}
Neural networks compute a sequence of feature vectors -- whose entries are called units -- through a sequence of transformations called layers. Common types of layers include affine/linear ones, that typically multiply by a weight matrix and add a bias vector, nonlinear layers, which typically apply an element-wise nonlinear function to their input (called an activation function), such as ReLU or tanh. Also increasingly common, and important to this work, are normalization layers, which normalize their inputs so that either 1) each entry/unit of the vector has sample mean 0 and variance 1 over the current minibatch of training data (called batch normalization \citep{ioffe2015batch}), or 2) the vector has sample mean 0 and variance 1 across its entries/units (called layer normalization \citep{ba2016layer}).

The weights of the neural networks are initialized to random small values, typically sampled from a zero-mean Gaussian, with two goals in mind: 1) to break symmetries in the network, allowing different hidden units to represent different features, and 2) to allow gradients to preserve norm, typically in expectations, making certain assumptions about the activation function in order to reason about the non-linearity of the system. Training is carried out by randomly sampling initial values of the parameters (typically the weights and biases of the affine layers) and then iteratively optimizing them with respect to an objective function. Objective functions are usually defined as the expected loss over the training set (with often additional regularization terms), and are estimated via iid sampling from the training set for the purposes of stochastic optimization. The most commonly used type of regularization is L2 regularization or Weight Decay (WD), which penalizes the squared norm of the parameter vector. Weight decay can also be incorporated directly into the weight update rule as opposed to being specified in the objective function to encourage the weights to be small in magnitude. The interplay between WD and BN has been extensively studied for a single task setting 
but remains poorly understood in the context of non-stationary data distribution.
Furthermore, we are specifically interested in studying the effects of the distribution shifts on the network's ability to retain the ability to continually learn. Due to the complexities of the optimization process, various factors contribute to how learning evolves especially in the presence of non-stationarity. In this work, we focus on the following phenomena: 1) \textit{unit saturation} (also referred to as gradient starvation \citep{pezeshki2021gradient}) measures the sufficiency of a gradient signal, the lack of which can prevent necessary features from being learned, 2) \textit{pre-activation distribution shift} measures the shift in the pre-activations of a unit which are often initialized to be constrained to zero-mean Gaussian as discussed before, 3) \textit{parameter growth} is aimed at measuring the growth in the magnitude of the norms of the parameters and features as large magnitudes can lead to instabilities in the optimization process, we use the notion of effective dimensionality of a network's representation to quantify this, and finally 4) \textit{loss landscape pathologies} are geared towards quantifying the sensitivity of the network's outputs to changes in its parameters, we use the notion of effective feature rank and its interplay with gradient similarity to measure this formally. We defer the formal details of how these are calculated to the respective sections in the paper later.

\section{Experiment details}
\label{appx:experiment-details}
\subsection{Network architectures}
\label{sec:architectures}
We consider three classes of network architecture: a fully-connected multilayer perceptron (MLP), for which we default to a width of 512 and depth of 4 in our evaluations; a convolutional network with $k$ convolutional layers followed by two fully connected layers, for which we default to depth four, 32 channels, and fully-connected hidden layer width of 256; we also consider a ResNet-18 architecture, which follows the standard architecture.

In all networks, we apply layer normalization before batch normalization if both are used at the same time. By default, we typically use layer normalization rather than batch normalization, although in many ablations we consider various combinations of one or both. 

\subsection{Continual supervised learning}
\label{appx:nonstat-benchmark-details}

Our continual supervised learning domain is constructed from a fixed image classification dataset: we have considered CIFAR-100, CIFAR-10, and MNIST and observe consistent results across all base datasets. We primarily use CIFAR-10 in our evaluations. Each continual classification problem is characterized by an input transformation and a label transformation. For input transformations, we use the identity transformation permutation of the image pixels. For label transformations, we permute classes (for example, all images with the label 5 will be re-assigned the label 2), and random label assignment, where each input is uniformly at random assigned a new label independent of its class in the underlying classification dataset.

We fix an iteration interval of 75000 optimizer steps, during which the network is trained using the adam optimizer. This number was selected as it was sufficient for all architectures to converge on most data transformations. We conducted a sweep over learning rates for the different architectures, settling on 1e-5 for the resnet, 1e-4 for the convolutional network, and 1e-5 for the MLP, to ensure that all networks could at least solve the single-task version of each label and target transformation. We then alternate between training the network, and applying a new random transformation to the dataset, for a fixed number of iterations, typically 20. 

\subsection{Contextual bandits}
\label{appx:contextual-bandit}

In the contextual bandit tasks, we train a reinforcement learning agent on a stationary environment using Q-learning with target networks as in DQN \citep{mnih2015human}. The environment is based on an image classification dataset (we consider CIFAR-10 in this paper), and is defined as an MDP with observations equal to the images in the dataset, actions equal to the number of classes, and a reward function which yields a value of $\alpha$ if the action taken in the state is equal to its label, and zero otherwise. The state then randomly transitions to a new image from the dataset. While it is not necessary to deploy a reinforcement learning algorithm in order to maximize the reward in this task, it provides a simple setting in which to study the loss of plasticity in reinforcement learning.

The DQN agents we train use target network update periods of either 500 or 5000 steps. We use a replay buffer of size 100,000, and have the agent follow a uniform random policy. In section~\ref{sec:mechanisms} we visualize results for convolutional neural networks, but we also run experiments on ResNets and MLPs. To evaluate plasticity in these agents, we take a network checkpoint, initialize a fresh optimizer, and train the network to fit a perturbation of its initial outputs on a subset of inputs from its replay buffer. Concretely, this means that if we let $\theta_T$ denote the parameter checkpoint we start from, the objective function of the probe task at step $k$ of the fine-tuning phase on some set of inputs $X$ will be equal to $\| f(\theta_T; X) - f(\theta_k; X) + \eta(X) \|^2$, where $\eta$ is a high-frequency function of the input $X$. We run this optimization procedure for a fixed number of steps (2000, selected as this was sufficient for most networks to solve the probe task at a random initialization). This probe task can be thought of as measuring the expected reduction in TD error if the network were to fit its Bellman targets with a random reward function, given some uniform prior over rewards.

\textbf{The `two-hot' trick:} we re-parameterize regression problems as classification problems by first setting some integer bound $M$ on the range of possible values our regression targets might take (for example, in a RL task with $\gamma=0.9$ and maximum reward $1$, we know that the maximum value of a state-action pair will be $M=10$). We then discretize the interval $[-M, M]$ to obtain a set of atoms $\{ -M, -M + 1, \dots, M-1, M\}$. The network's output is a $2M + 1$-dimensional vector of logits to a softmax distribution on this set. Given a regression target $c$, we construct a probability distribution with support on $\lfloor c\rfloor$ and $\lceil c \rceil$, with probabilities $P(\lfloor c \rfloor) = \lceil c \rceil - c$, and $P(\lceil c \rceil) = 1 - P(\lfloor c \rfloor)$. We then minimize the cross-entropy loss between this distribution and the softmax distribution output by the network.

\subsection{Natural non-stationarities}
\label{appx:wilds-atari}

\textbf{WiLDS}: The WiLDS benchmark consists of a collection of datasets which can be partitioned to induce distribution shifts. In keeping with prior analyses of image classification, the dataset used in \cref{fig:natural-nonstats} is the \texttt{iwildcam} dataset, which consists of photos of animals taken in a variety of locations and times of day. To generate a series of distribution shifts, we train a network only on data collected from a single location; at fixed intervals, we change the location used to generate the training data, and continue training on this new location. We use 10 locations and repeat this process 20 times.

\textbf{Arcade learning environment:} we also study deep reinforcement learning agents on the arcade learning environment. While we only consider single-environment tasks, we note that the RL process itself introduces a number of rich forms of nonstationarity, in both the distribution of inputs and the learning targets. We study networks trained with the C51 algorithm, a distributional approach which approximates the probability distribution of returns from a given state-action pair, rather than their expected value, using a categorical distribution on 51 atoms. The learner then minimizes the KL divergence between its predicted distribution and the \textit{distributional Bellman target}. The precise details of this algorithm can be found in \citet{bellemare2017distributional}.

Its relevance for us is that, similarly to the two-hot trick, it leverages the scale-invariant cross-entropy loss in order to solve what is ultimately a regression problem. This allows us to verify that, as in the contextual bandits discussed previously, combining layer normalization and scale-invariant losses can allow the network to maintain its ability to adapt to changes in its prediction targets. We further note that, unlike the two-hot loss, distributional objectives are less likely to struggle with saturated softmax outputs that proved to be a challenge in the contextual bandit tasks because the distributional Bellman targets will typically exhibit nontrivial entropy.

\section{Neural tangent kernel for linearized networks}

We provide some additional insight into why linearization can lead to pathological learning dynamics by considering the empirical neural tangent kernel obtained by a linear network. We first note that in our case a \textit{linearized} network will have identical local dynamics to a \textit{linear} network of the form $\prod_{l=1}^L M_{\dot{\phi}}^l W^l $, where $M_{\dot{\phi}}^l$ denotes the matrix of slopes of the nonlinearity $\phi$ for each feature. In a ReLU network with all positive slopes, $M_{\dot{\phi}}^l = Id$ for each layer $l$, but in a network with dormant neurons one would have some zero entries on the diagonal of $M_{\dot{\phi}}$. Similarly, a leaky ReLU network would have some slope values of $-\alpha$, where $-\alpha$ denotes the slope assigned to negative preactivation values. Thus in general, we have
\begin{equation}
    f_\theta(\bx) = \bigg ( \prod_{l=1}^L M_{\dot{\phi}} W^l \bigg ) \bx \; .
\end{equation}

For simplicity, we assume $f_\theta : \mathbb{R}^{d_{in}} \rightarrow \mathbb{R}$, with $\theta = \oplus_{k=1}^L \mathrm{vec}(W_k)$ (i.e. there are no bias terms in the network). This results in the following per-layer Jacobian:
\begin{equation}
\nabla_{W^k} f_\theta(\bx) = \prod_{l=k+1}^{L} M_{\dot{\phi}} W^l \bx^\top \bigg ( \prod_{l=1}^{k-1} M_{\dot{\phi}} W^l \bigg )^\top
\end{equation}
If we vectorize $W^k$, we then obtain the following linear function of $\bx$:
\begin{align*}
   \mathrm{vec}( \nabla_{W^k} f_\theta(\bx) )_{ij}& = \prod_{l=k+1}^{L} M_{\dot{\phi}} W^l \bx^\top \bigg ( \prod_{l=1}^{k-1} M_{\dot{\phi}} W^l \bigg )^\top [i, j] \\
   &= (\prod_{l=k+1}^L M_{\dot{\phi}} W^l  )[i] ( \prod_{l=1}^{k-1} W^l \bx)[j] \\
   \intertext{letting $M_{>k} =\prod_{l=k+1}^L M_{\dot{\phi}} W^l  $}
   &= M_{>k}[i] \sum_{i' = 1}^{d_{in}}\bigg ( \prod_{l=1}^{k-1}M_{\dot{\phi}} W^l \bigg )[j, i'] \bx_{i'} \\
   \intertext{ and analogously $M_{<k} =\prod_{l=1}^{k-1}M_{\dot{\phi}} W^l  $}
   &= M_{>k}[i] \sum_{i' = 1}^{d_{in}} M_{< k}[j, i'] \bx_{i'} \\
   \implies \mathrm{vec}( \nabla_{W^k} f_\theta(\bx) ) &= A_k \bx \text{ where } A_k[ij, i'] = M_{>k}[i] M_{<k}[j, i']
\end{align*}

We can then write up the empirical neural tangent kernel for the entire set of parameters $\theta = W_1 \oplus \dots \oplus W_L$ as 
\begin{align*}
    K_{\theta}(\bx, \by) &= \sum_{k=1}^L \langle( \mathrm{vec}(\nabla_{W^k}(f_\theta)(\bx)),  \mathrm{vec}(\nabla_{W^k}(f_\theta)(\by)) \rangle \\
    &= \sum_{k=1}^L \bx^\top ( A_k^\top A_k) \by = \bx^\top (\sum_{k=1}^L A_k^\top A_k) \by
    \intertext{ and consequently}
    K_{\theta}(\bX, \bX ) &= \bX^\top (\sum A_k ^\top A_k) \bX \\
    \implies \mathrm{rank}(K_{\theta}(\bX, \bX )) & \leq \mathrm{rk}(\bX)
\end{align*}

As a result, we see that the local training dynamics of a linearized network have bounded rank, in the best case equal to the input dimension but possibly much lower, particularly if the training samples lie in a low-dimensional subspace of the input space. For example, if a network has many dead ReLU units then it may come to be that the matrices $A_k$ end up bottlenecking the rank of the NTK more so than the dimension of the training samples $\bX$.

\section{Additional Analysis}
In this section, we include additional ablations and finer-grained analyses relating to the results in the main paper.
\subsection{Visualizing learning curves}

\begin{figure}
    \centering
    \includegraphics[width=\linewidth]{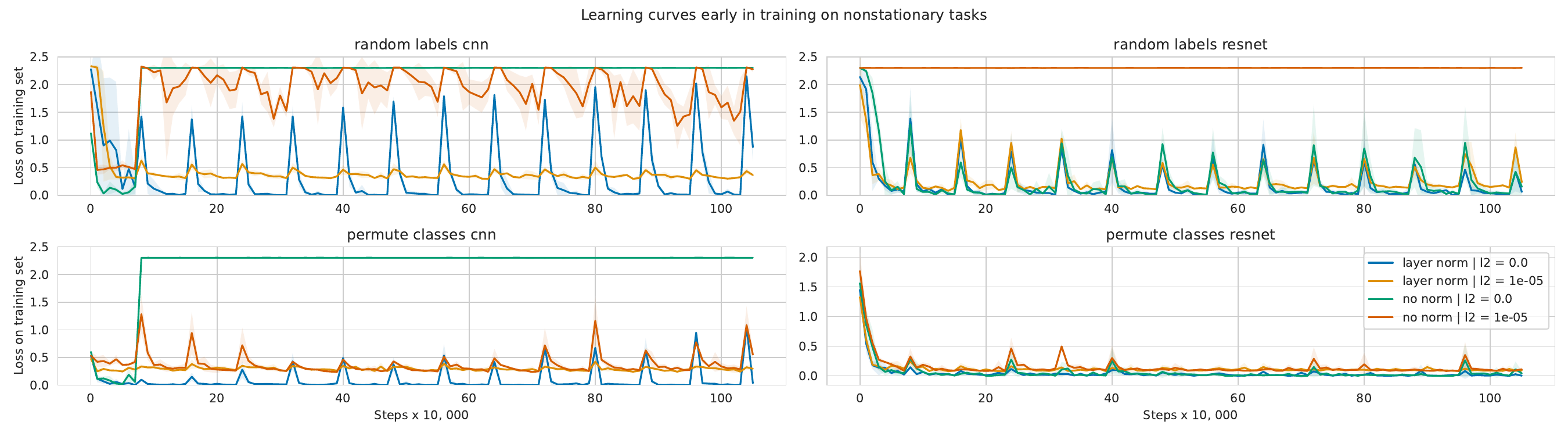}
    \includegraphics[width=\linewidth]{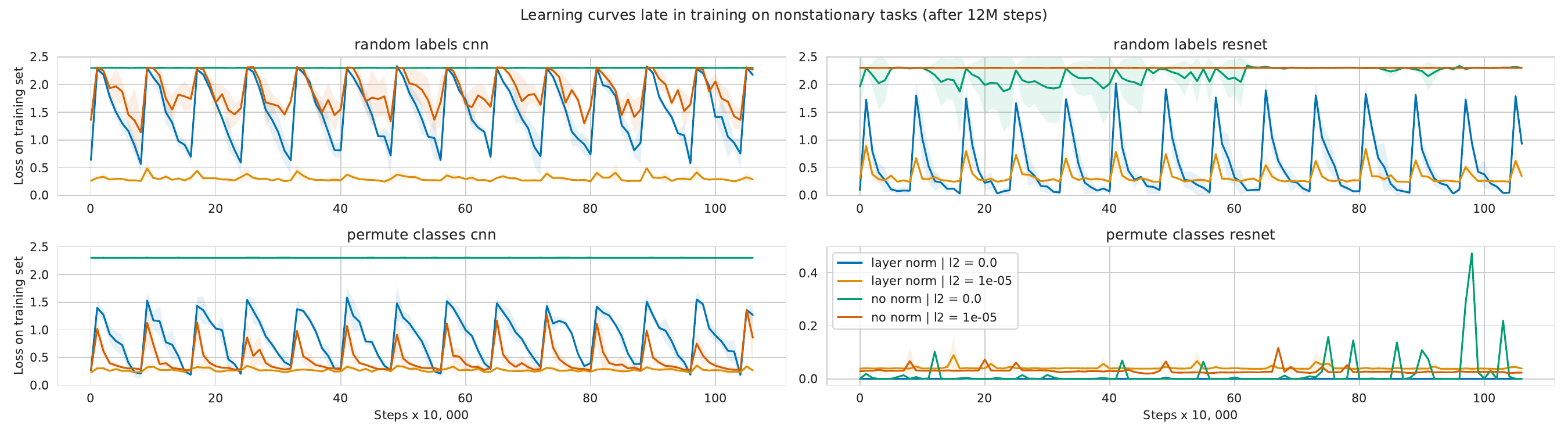}
    \caption{Learning curves early and late in a nonstationary training evaluation.}
    \label{fig:in_depth}
\end{figure}

While most of the figures in this paper only show accuracy at the end of a task iteration, we also visualize what the loss curves look like at a more fine-grained level in \cref{fig:in_depth}. Note that in many cases, declines in plasticity later in training often correspond to learning curves becoming shallower, rather than to higher plateaus.

\subsection{Normalization layer components}
\begin{figure}
    \centering
    \includegraphics[width=\linewidth]{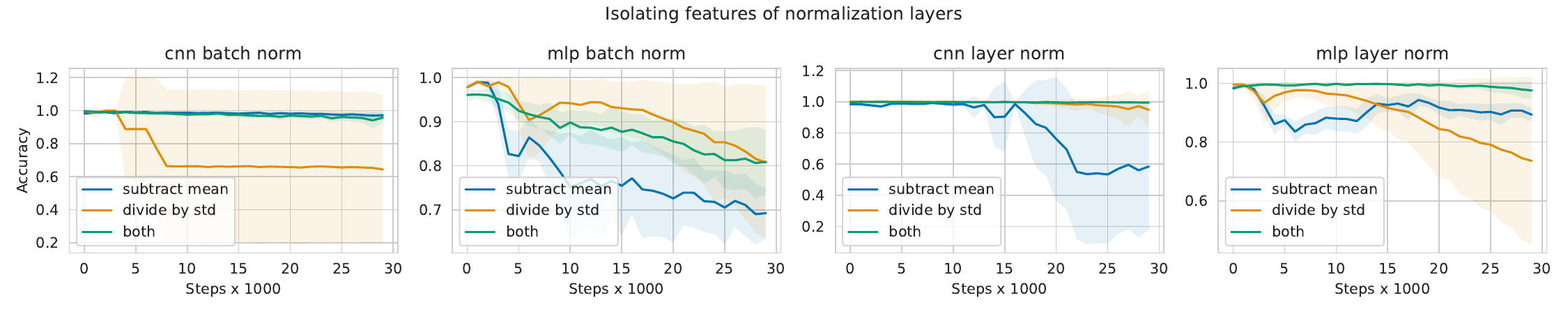}
    \caption{Ablating role of mean subtraction and normalization by standard deviation.}
    \label{fig:appx-normalization}
\end{figure}

In \cref{fig:appx-normalization} we ablate the roles of the different components of normalization layers. In particular we are interested in disentangling the relative importance of centering the preactivation distribution about zero, and rescaling to unit standard deviation. We find that most of the performance gain from normalization layers can be attributed to the second mechanism, although the relative effect sizes of each does vary across normalization dimension, network architecture, and dataset.

\subsection{Interaction between nonlinearity and normalization}
\begin{figure}
    \centering
    \includegraphics[width=\linewidth]{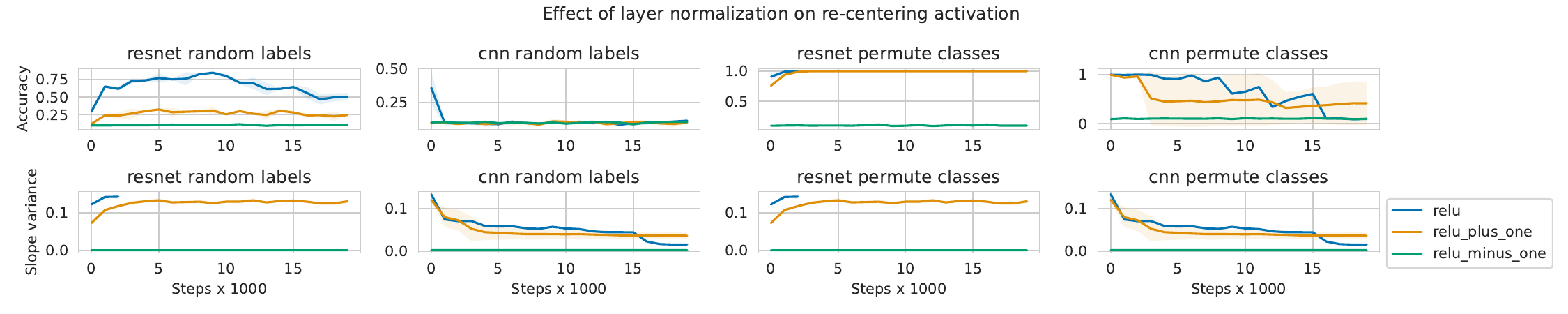}
    \includegraphics[width=\linewidth]{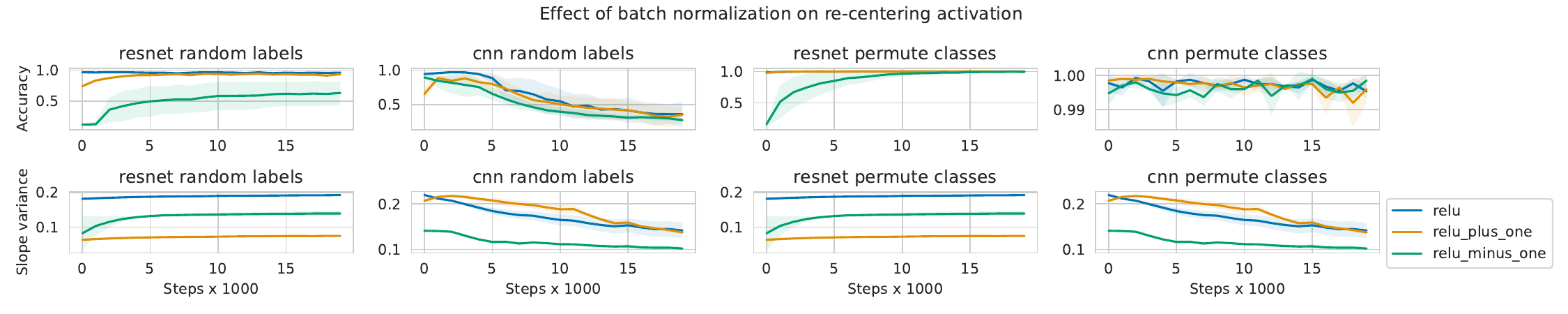}
    \includegraphics[width=\linewidth]{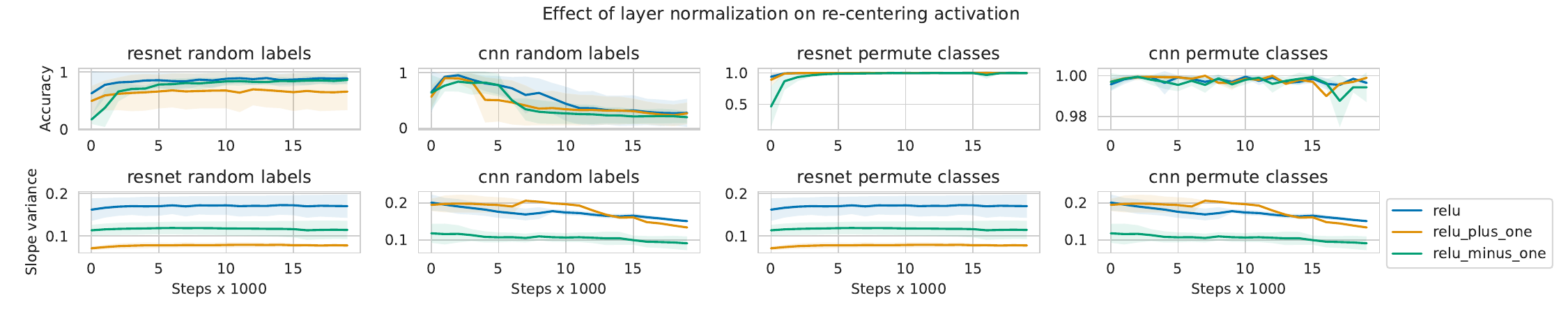}
    \caption{Effect of re-centering ReLU units on plasticity and expressivity.  }
    \label{fig:relu_variants}
\end{figure}

In \cref{fig:relu_variants}, we evaluate the importance of normalization layers centering their normalization precisely around the region where most nonlinearities exhibit the greatest effect (i.e. zero in the case of ReLU, GeLU, tanh, and leaky ReLU networks). In particular, we see whether adding normalization layers still exhibits a similar benefit to plasticity when the normalization centering occurs away from the `receptive field' of the nonlinearity. We study this in the context of ReLU units where we offset the input by a fixed value, either $+1$ or $-1$. Intriguingly, while we do observe a reduced variance in the slope of the nonlinearity when an off-center activation is used, this does not always result in reduced performance. Indeed, in many cases the normalized networks outperform an unnormalized network on the tasks despite reduced variability in the behaviour of the nonlinearity. This observation suggests that the role of normalization layers in maintaining plasticity may have less to do with the precise range in which they keep the activations, and more to do with the stability of that range.

\subsection{Effect of depth and width}
\begin{figure}
    \centering
    \includegraphics[width=0.99\linewidth]{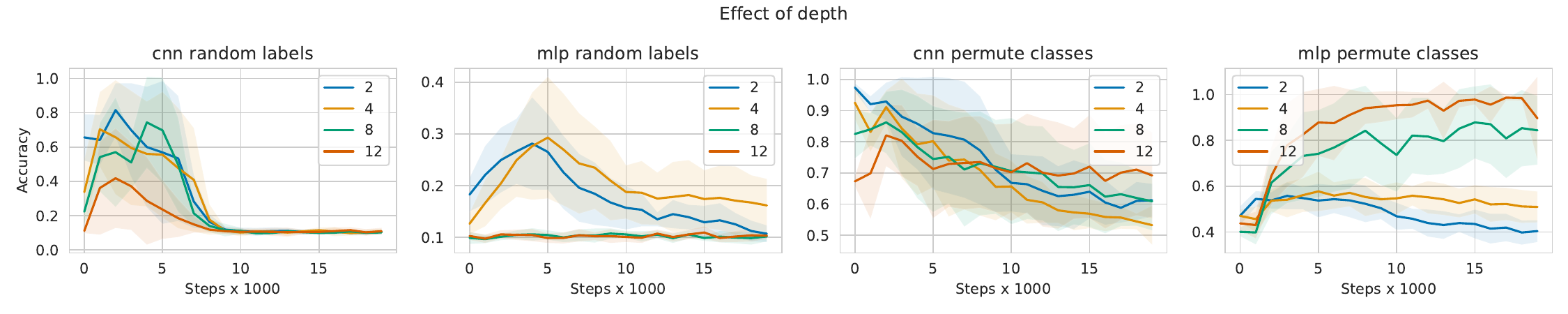}
    \includegraphics[width=0.99\linewidth]{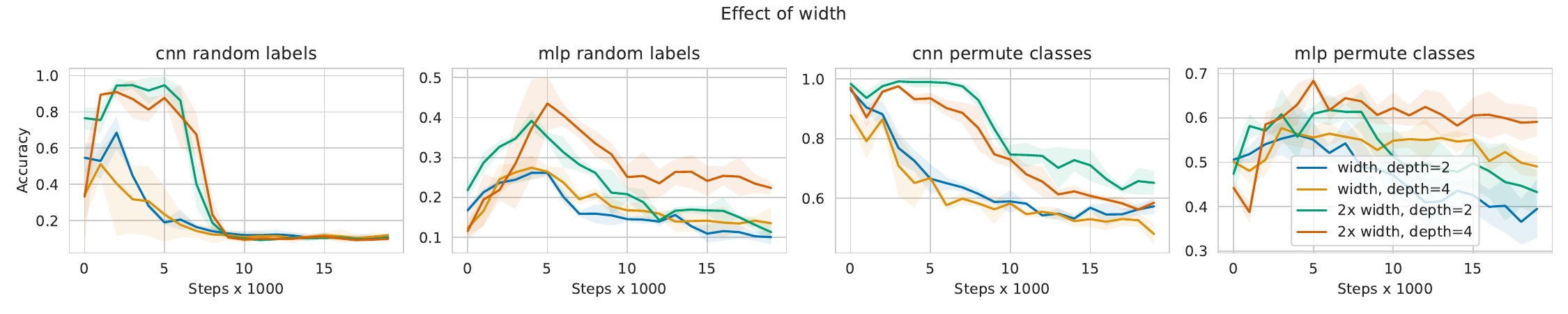}
    \caption{\textbf{Top:} relationship between depth and plasticity. We see that networks benefit from depth in the permute-classes task but not in the random-labels task, where increased depth without additional hyperparameter tuning reduces performance and accelerates the loss of plasticity. \textbf{Bottom:} relationship between width and plasticity. Across depths, increasing width is beneficial. }
    \label{fig:appx-depth}
\end{figure}

In \cref{fig:appx-depth} we investigate the relationship between network size and robustness to nonstationarity. We do not perform any additional hyperparameter sweeps for the different depths. Perhaps surprisingly, we find that in spite of using a learning rate that was tuned for a depth 4 network, using a depth of up to 12 results in improved performance in the permute-classes setting. In contrast, the deeper networks exhibit greater difficulty even in the first iteration of the random labels task. In contrast, and in agreement with the observations of \citet{lyle2023understanding}, we do see consistent improvements from increasing width.

\section{Additional results}
\label{sec:rebuttal}
\subsection{Output scale and plasticity}
\label{sec:scale-plasticity}
\begin{figure}
    \centering
    \includegraphics[width=\linewidth]{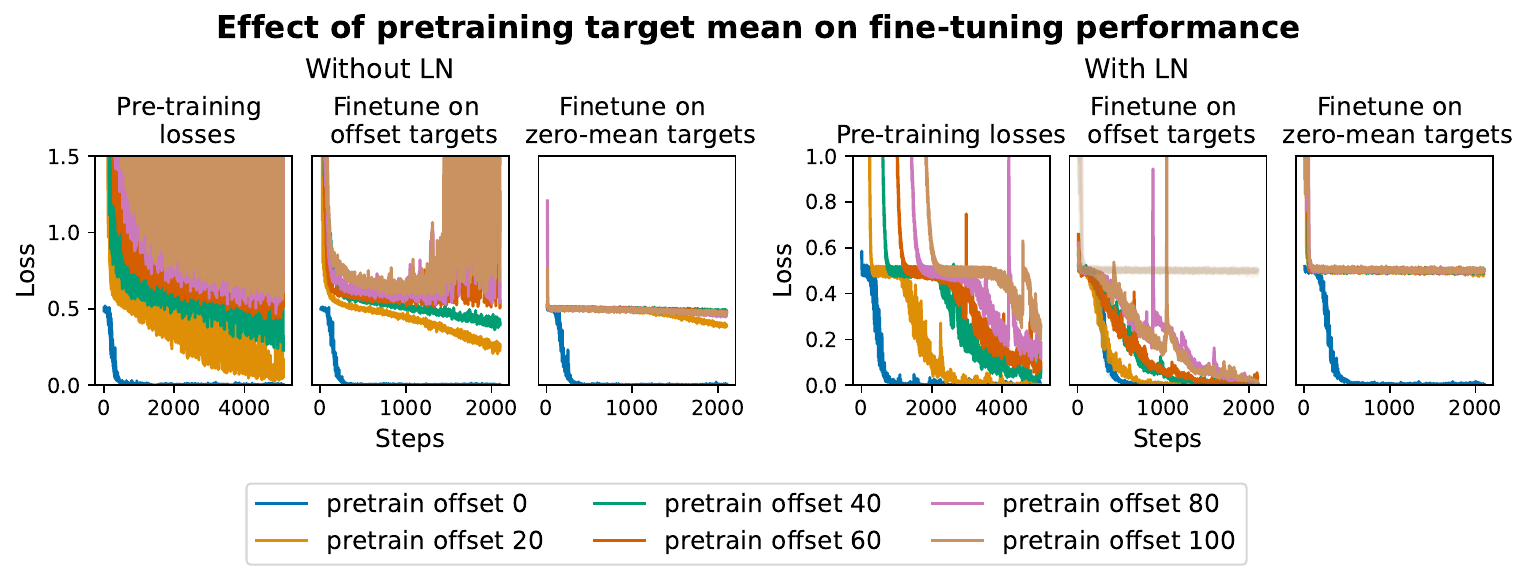}
    \caption{Detailed learning curves for networks with and without layer normalization when trained on regression tasks with varying pre-training offset means.}
    \label{fig:detailed-offset-curves}
\end{figure}

\begin{figure}
    \centering
    \includegraphics[width=0.42\linewidth]{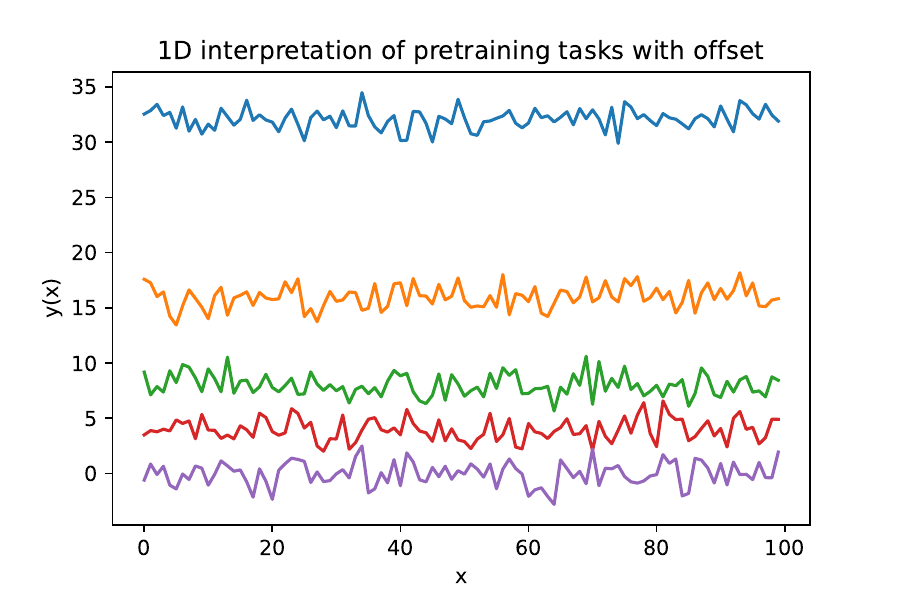}
    \includegraphics[width=0.57\linewidth]{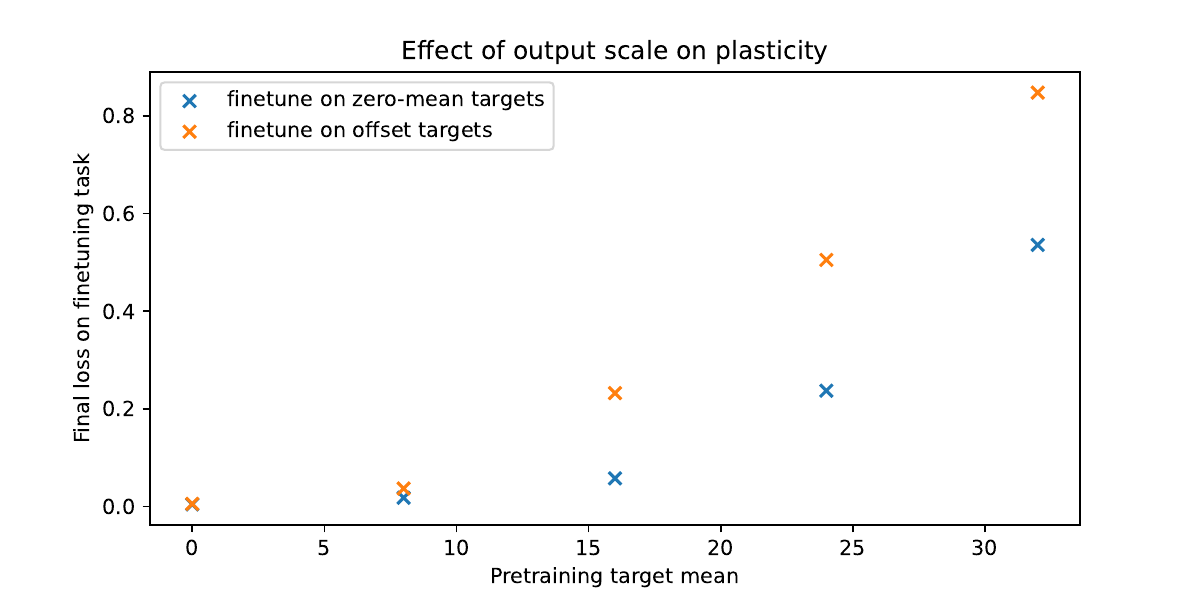}
    \caption{Left: Visualization of a 1D analogue the regression targets used for pre-training (actual targets are constructed over an image dataset to allow for more realistic network architectures). Right: ``dose-response" curve for pretraining target offset scale vs final performance on fine-tuning task with either zero offset or the same offset as the pretraining data. In both cases, we see similar reductions in fine-tuning performance as a function of the pretraining offset.}
    \label{fig:regression-target-viz}
\end{figure}

We illustrate the effect of increasing $\gamma$ on the plasticity of deep Q-learning agents in Figure~\ref{fig:param-growth}. While increasing $\gamma$ increases the scale of the TD targets, it also increases the degree of bootstrapping performed by the network, and it is difficult to disentangle these potentially competing mechanisms from the figure alone. To provide additional support for our claim that output scale can damage the plasticity of neural networks independent of bootstrapping, we distill the target phenomenon down to its most essential components. We perform standard mean squared error minimization on a set of random targets to which we add a fixed bias term. In our experiments, we use the CIFAR-10 dataset as inputs, and construct challenging regression targets by applying a high-frequency sinusoidal transformation to the output of a randomly initialized neural network. Specifically, for $M=10^5$, we have
$f_{\mathrm{target}}(\mathbf{x}) = \sin (M f_{\theta_{\mathrm{rand}}}(\mathbf{x})) + b$, where $b$ is some fixed bias. A one-dimensional interpretation of the regression targets for different values of $b$ is provided in Figure~\ref{fig:regression-target-viz}. We provide more detailed learning curves for networks both with and without layer norm in the architecture from Figure~\ref{fig:param-growth} in Figure~\ref{fig:detailed-offset-curves}.

\begin{figure}
    \centering
    \includegraphics[width=1.0\linewidth]{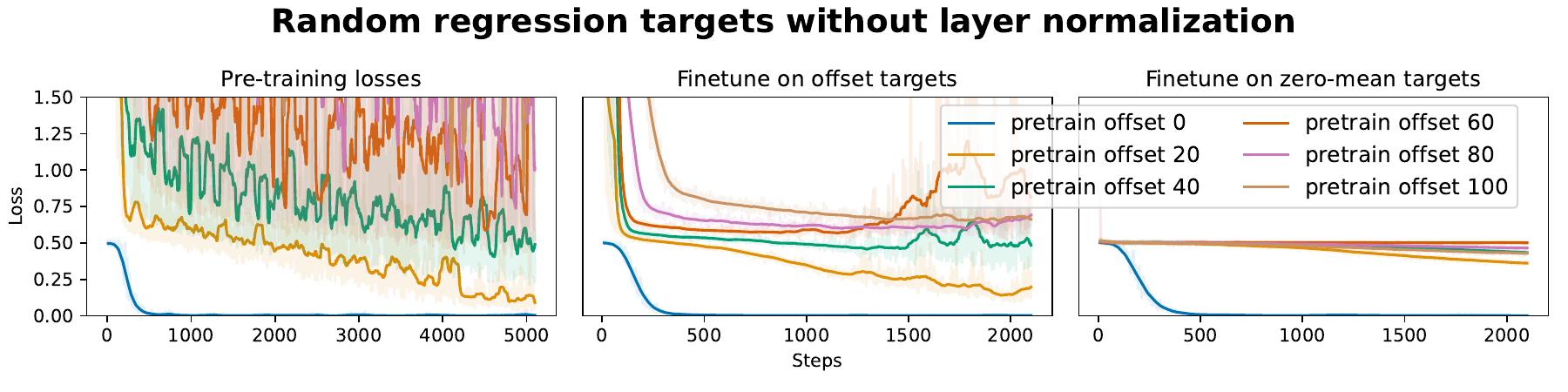}
    \caption{Effect of pretraining bias magnitude on finetuning accuracy. We omit the first 1000 steps of the fine-tuning period to allow for a more informative scale.}
    \label{fig:regression-finetuning-scale}
\end{figure}

In our experimental framework, we pretrain a different convolutional network (the same CNN architecture as is used in the ``standard" RL training regime) on a target function defined as above for each $b \in \{0, 8, 16, 32 \}$. We then construct a new set of targets, which use a different random target network initialization and have either bias equal to the pretraining bias or $b=0$. In both cases, we see that the network pretrained on $b=0$ is better able to reduce the loss on the finetuning targets. We further see a monotone relationship between the size of the pretraining offset and the final loss of the network. Averaged learning curves over three random target seeds are shown in Figure~\ref{fig:regression-finetuning-scale}, and averaged final losses (we average both over seeds and over the final 10 SGD iterates to reduce noise) are shown in Figure~\ref{fig:regression-target-viz}. We run the experiment on a subset of 10,000 images from the underlying dataset and use a batch size of 512; we use the Adam optimizer with learning rate 0.001.

Looking into the singular value decomposition of the penultimate-layer features of the network, obtained by performing SVD on a $n \times d$ matrix given by feature embeddings for $n$ randomly sampled inputs with $n=128$ in this case, gives some insight into what is driving the optimization difficulties in networks trained to minimize a regression loss on a target of the form $f(x) = 100 + \epsilon(x)$. In networks which don't include layer normalization, we see an immense increase in the maximal singular value as the pretraining target offset grows (ranging from $O(10^3)$ for mean-zero targets to $O(10^8)$ for mean-100 targets). Layer normalization constrains the maximum norm of the features in networks which incorporate it and thus the maximum singular value of the feature matrix, however even in these networks we see a significant decline in the magnitudes of lower-order singular values relative to the largest one.

\begin{figure}
    \centering
    \includegraphics[width=\linewidth]{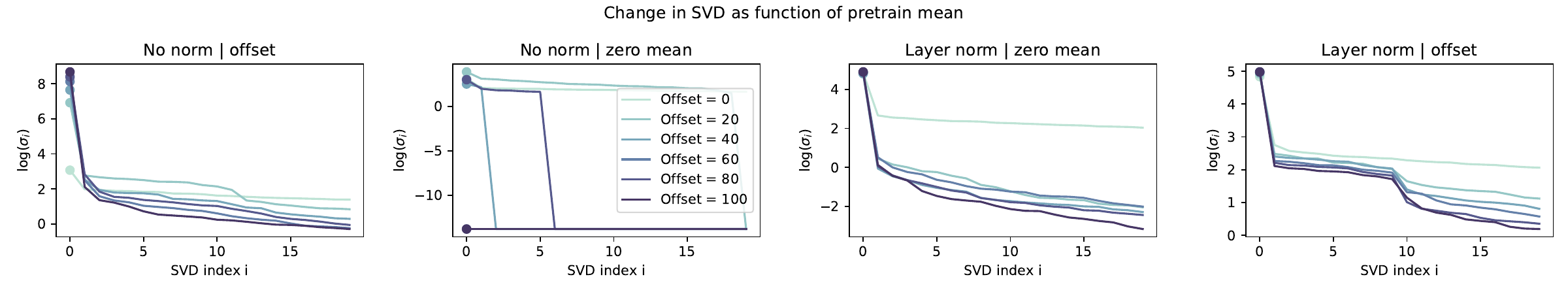}
    \includegraphics[width=\linewidth]{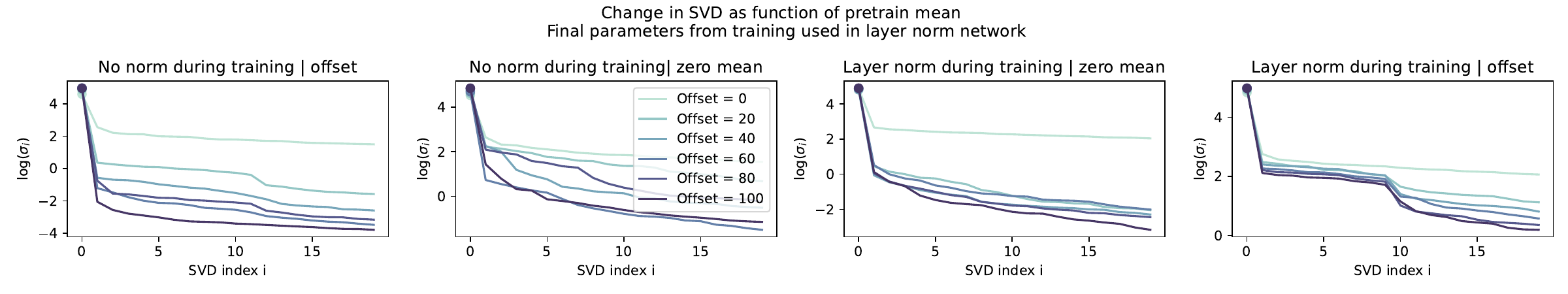}
    \includegraphics[width=\linewidth]{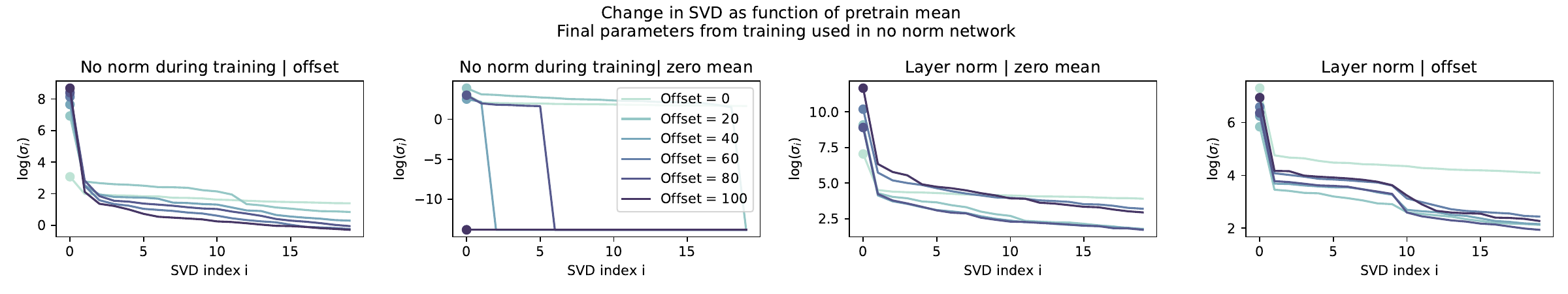}
    \caption{Distribution of top singular values as a function of index (note log scale). Magnitude of the top singular value grows with pretraining offset norm and remains high even after fine-tuning on We observe that networks with layer normalization have less extreme ill-conditioning of their features than networks without layer normalization.}
    \label{fig:my_label}
\end{figure}

\begin{figure}
    \centering
    \includegraphics[width=\linewidth]{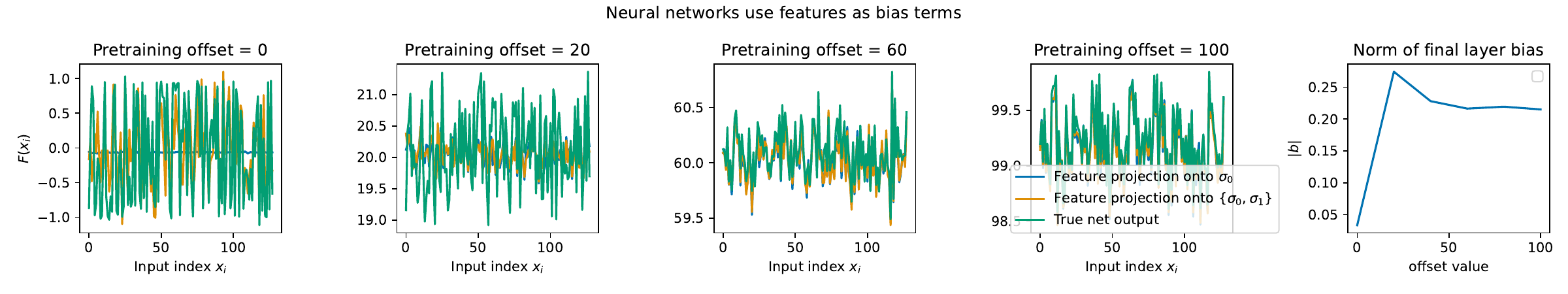}
    \includegraphics[width=\linewidth]{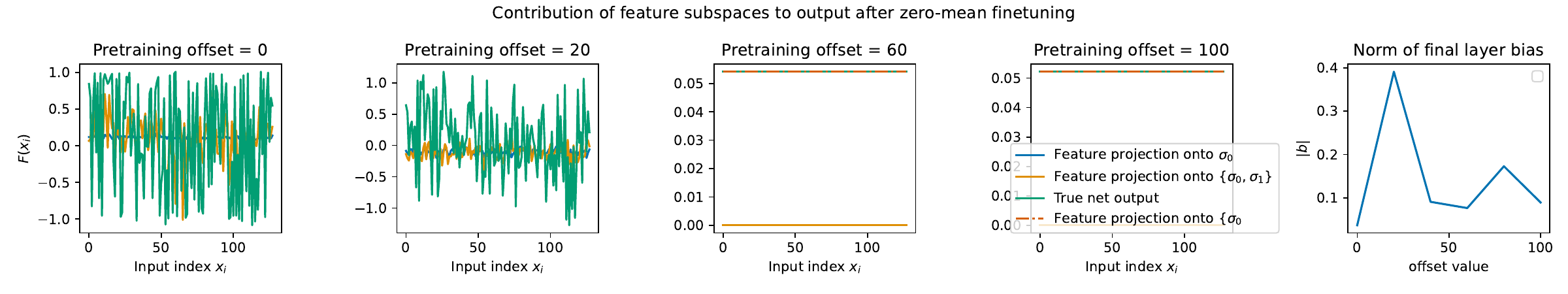}
    \caption{Visualization of contribution of principal components of penultimate layer features to the final network output, in contrast with the contribution of the final-layer bias term.}
    \label{fig:svd_bias}
\end{figure}

And what exactly does this largest singular value correspond to in feature space? As we can see in Figure~\ref{fig:svd_bias}, this is the dimension in which the network is encoding the bias term of the targets. First, we note that a neural network with nonzero bias weights in its final output layer should in theory be able to represent functions with arbitrarily large means by increasing the corresponding bias term; however, this does not happen in practice. We observe in the rightmost plot of Figure~\ref{fig:svd_bias} that the norm of the bias weights in these networks does not exhibit a monotone trend as the pretraining target offset grows. This is particularly striking in contrast to the trends observed for the feature embeddings, which grew not only monotonically but by several orders of magnitude. 

Though their bias terms are small, these networks are nonetheless outputting functions whose mean corresponds to that of the pretraining targets, which means that a large constant function must emerge from somewhere. As it turns out, rather than encoding this mean in the bias weights, the networks have instead learned to produce features which can be used as effective bias terms, by encoding the bias term into a single-dimensional subspace of the feature space with which feature embeddings have roughly constant dot product. To study the contribution of this single-dimensional  subspace of the embedding space to the network's output, we project the feature matrix onto that subspace and then apply the final layer weights to the projected matrix, visualizing the resulting projected outputs for 128 randomly sampled inputs as a (highly disordered) line plot. Strikingly, these visualizations make clear that the subspace encoded by the principal singular value is being used by these networks as essentially a bias term, to which perturbations arising from other subspaces are added. As target magnitude increases, so does the deviation from a perfectly constant function along with the relative contribution of lower singular vectors. Notably, in the networks which are fine-tuned on targets with mean zero, the corresponding dimension of feature space contributes almost zero to the final fine-tuned output, suggesting that the network is using this dimension as an offset term and overwrites it when this offset is no longer needed (though in the case of larger magnitudes this is accidentally achieved by saturating units, as we can see in the second row of Figure~\ref{fig:svd_bias} that the network output is characterized entirely by the bias).

As a result, we see that enormous singular values and ill-conditioning of the feature embeddings is a direct consequence of outputting a large, near-constant value. This occurs in its most extreme form in networks trained without layer normalization: in networks with layer normalization, this ill-conditioning effect is somewhat mitigated but nonetheless still occurs. Although in theory a network could increase the magnitude of the bias term in its output layer to achieve a larger output mean, in practice networks encode a `bias' term in their features instead. 

\begin{figure}
    \centering
    \includegraphics[width=0.7\linewidth]{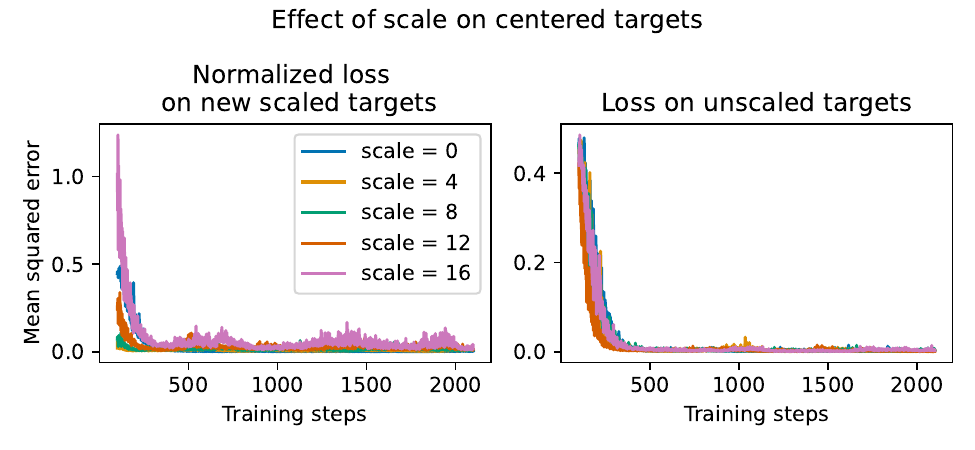}
    \caption{We contrast our previous analysis of offset targets with targets which are centered and scaled. Similar to Figure~\ref{fig:regression-target-viz}, we construct a pre-training task of the form $c + \alpha f_\theta(\bx) + \sigma(\bx)$, where $f_\theta$ is the output of a randomly initialized network and $\sigma$ is a high-frequency function. Rather than scaling $c$, we instead scale the smooth part of the function by a factor $\alpha$. In contrast to the case of scaled $c$, we don't see an effect of pre-training scale on finetuning loss on $\alpha = 0$. While we do see an increased loss when finetuning on targets with larger $\alpha$, this increase is in line with the natural increase in mean squared error obtained by scaling targets and predictions. }
    \label{fig:centered-and-scaled}
\end{figure}

In order to disentangle effect of output magnitude from output mean, we run an additional experiment involving mean zero targets that are scaled by different factors. The results from this experiment are given in Figure \ref{fig:centered-and-scaled}. From these we see that it's specifically a large target mean, and not necessarily large targets in general, that lead to plasticity loss.

\subsection{Uneven parameter norm growth}
\label{sec:unequal-param-growth}

\begin{figure}
    \centering
    \includegraphics[height=4cm]{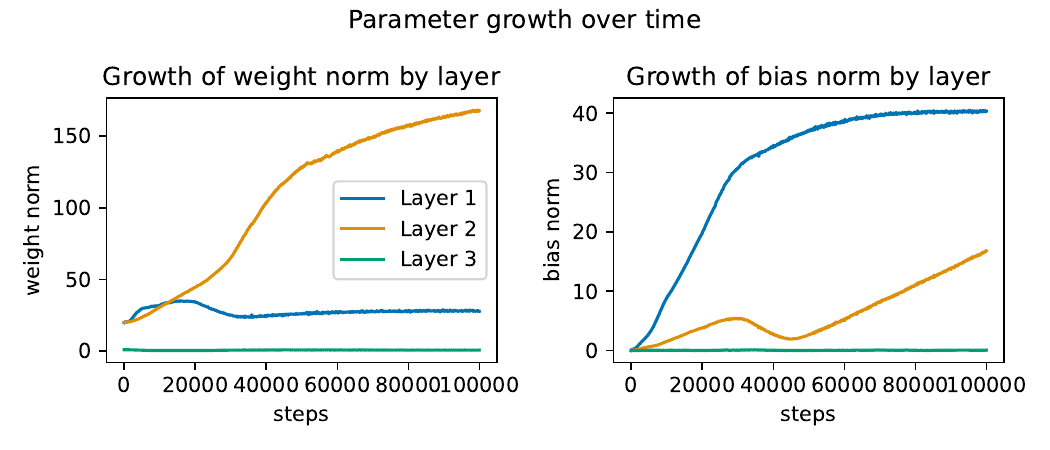}
    \includegraphics[height=4cm]{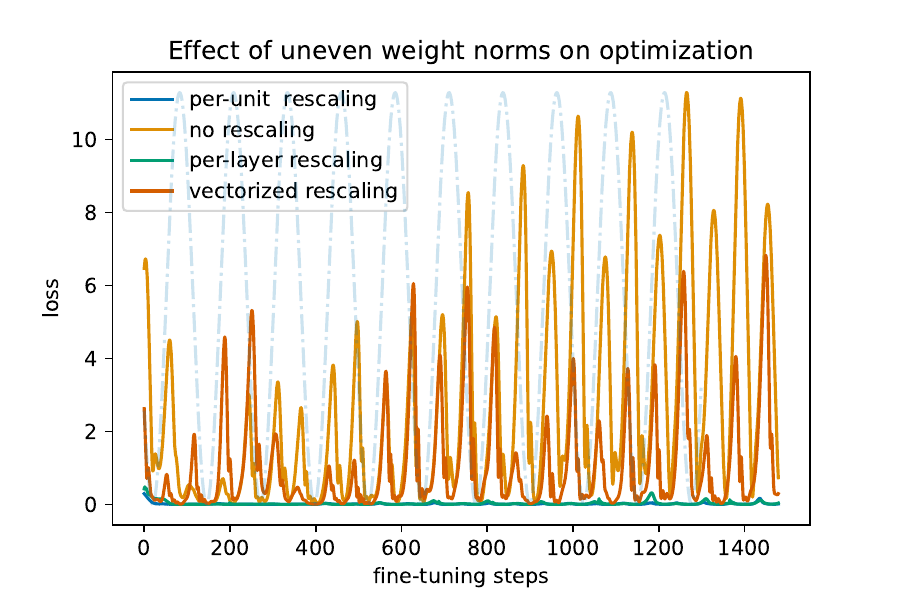}
    \caption{Parameter norm growth of different layers in an MLP trained on nonstationary regression}
    \label{fig:param-growth-unequal}
\end{figure}

We illustrate that even in a small MLP, differential scaling of the input, hidden, and output layers can result in markedly different trends in the growth of the parameters in these layers. The network is trained on an extremely simple task: we fix some randomly generated inputs and targets $\bx, \by$, and then regress the network on $\by + \sin (t/20)$ where $t$ is the number of optimization steps taken so far. We also observe that this unequal growth can be accompanied by optimization difficulties if these networks are then trained on a new task. 

\subsection{Death of a ReLU neuron}

Figure~\ref{fig:dead-and-zombies} illustrates at a high level how early gradients are biased towards either reducing or increasing pre-activation magnitudes over all inputs. It is natural to then conclude that, because updates immediately after a task change can be extremely large due to out-of-date second-order estimates in adaptive optimizers such as Adam, this bias will have the effect of potentially killing off many units. In this section, we provide a step-by-step illustration of how units die off, and why sudden task changes are particularly hazardous for ReLU activations. We rely on the visualizations in Figure~\ref{fig:unit-death-norm}. Concretely, the following four factors drive the death of ReLU units:

\begin{figure}
    \centering
    \includegraphics[width=\linewidth]{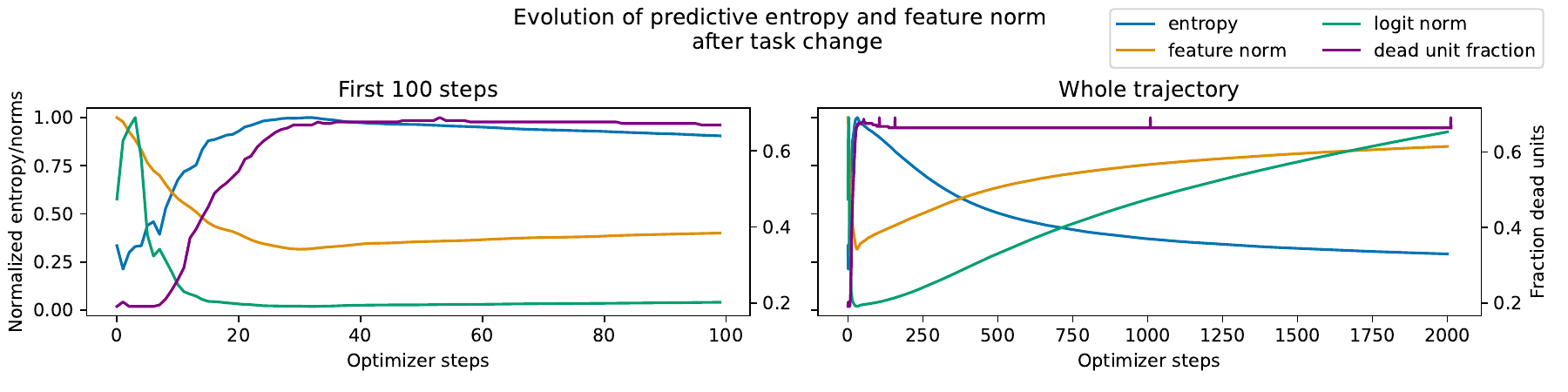}
    \includegraphics[width=\linewidth]{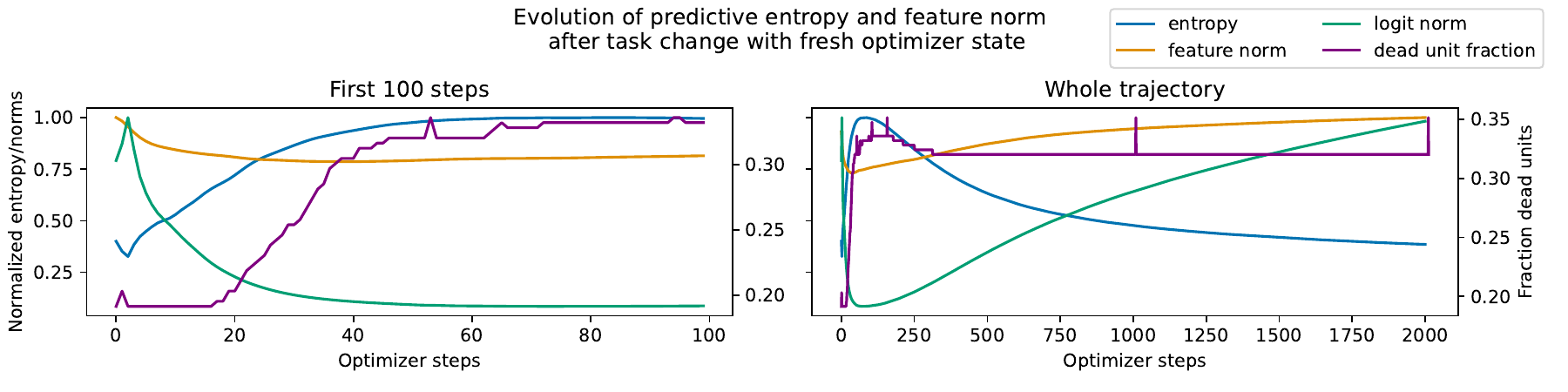}
    \caption{Evolution of logit norm, feature norm, logit norm, and dead unit fraction in a one-hidden-layer MLP trained to memorize one set of random MNIST labels, and then fine-tuned on a second set. Results show early and total training dynamics on the second set of random labels. Top: initial optimizer state and parameters taken from the final step of the previous task. Bottom: resets the adam optimizer before continuing training on the second set of random labels. Note the smaller right-hand-side y-axis scale in the bottom figure, corresponding to half the number of dead units.}
    \label{fig:unit-death-norm}
\end{figure}

\begin{enumerate}
    \item Immediately after a task change, the network aims to increase the predictive entropy of its outputs. This can be seen in the increase in output entropy immediately following the task change, along with the corresponding decrease in logit norm over the first hundred optimizer steps.
    \item There are two ways to reduce the magnitude of the incorrect class probability: one can either reduce the norm of the final-layer weight associated with that logit, or one can reduce the norm of the incoming features. As we can see, the network tends to decrease the norm of the incoming features. 
    \item Gradients which reduce the norm of the features have the effect of reducing the pre-activation value for most inputs, since negative pre-activations do not contribute to the feature norm due to the ReLU. We see that in many units the gradient has a negative dot product with all inputs. 
    \item If the optimizer state is not reset, then outdated second-order estimates cause large update steps. Large update steps in a direction which push down pre-activations quickly results in unit death, with all pre-activations becoming negative within a handful of steps. 
\end{enumerate}

\subsection{Detailed Atari results}
\label{appx:detailed-atari}
\begin{figure}
    \centering
    \includegraphics[width=0.4\linewidth]{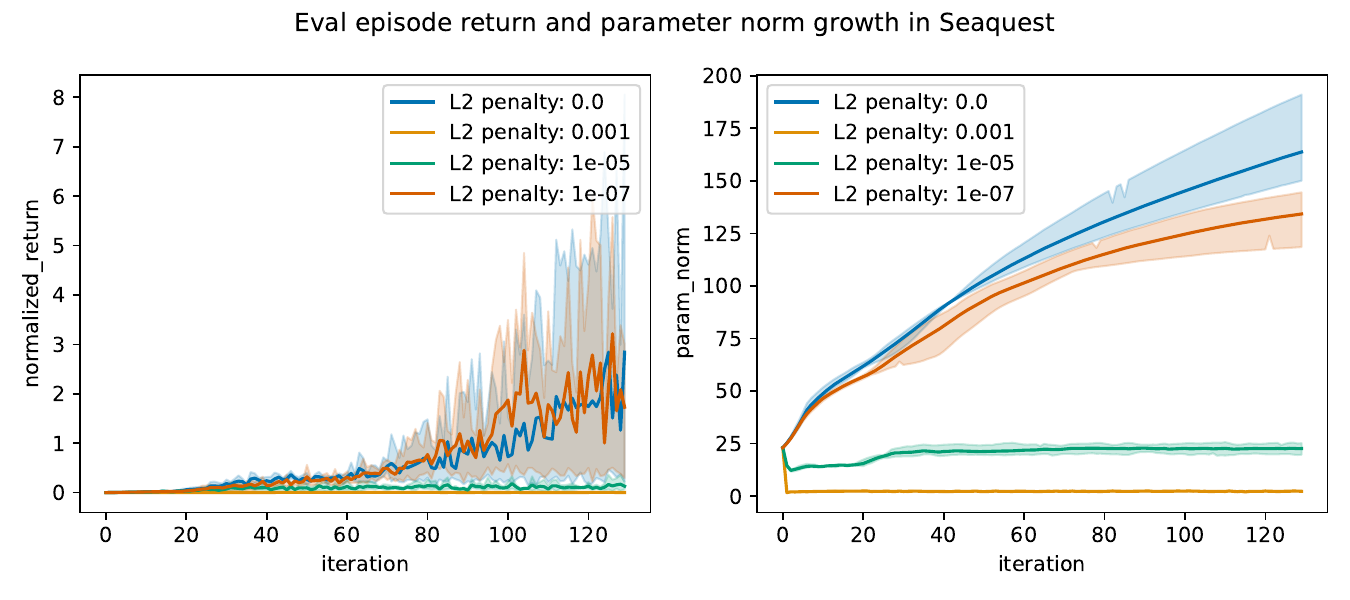}
    \includegraphics[width=0.58\linewidth]{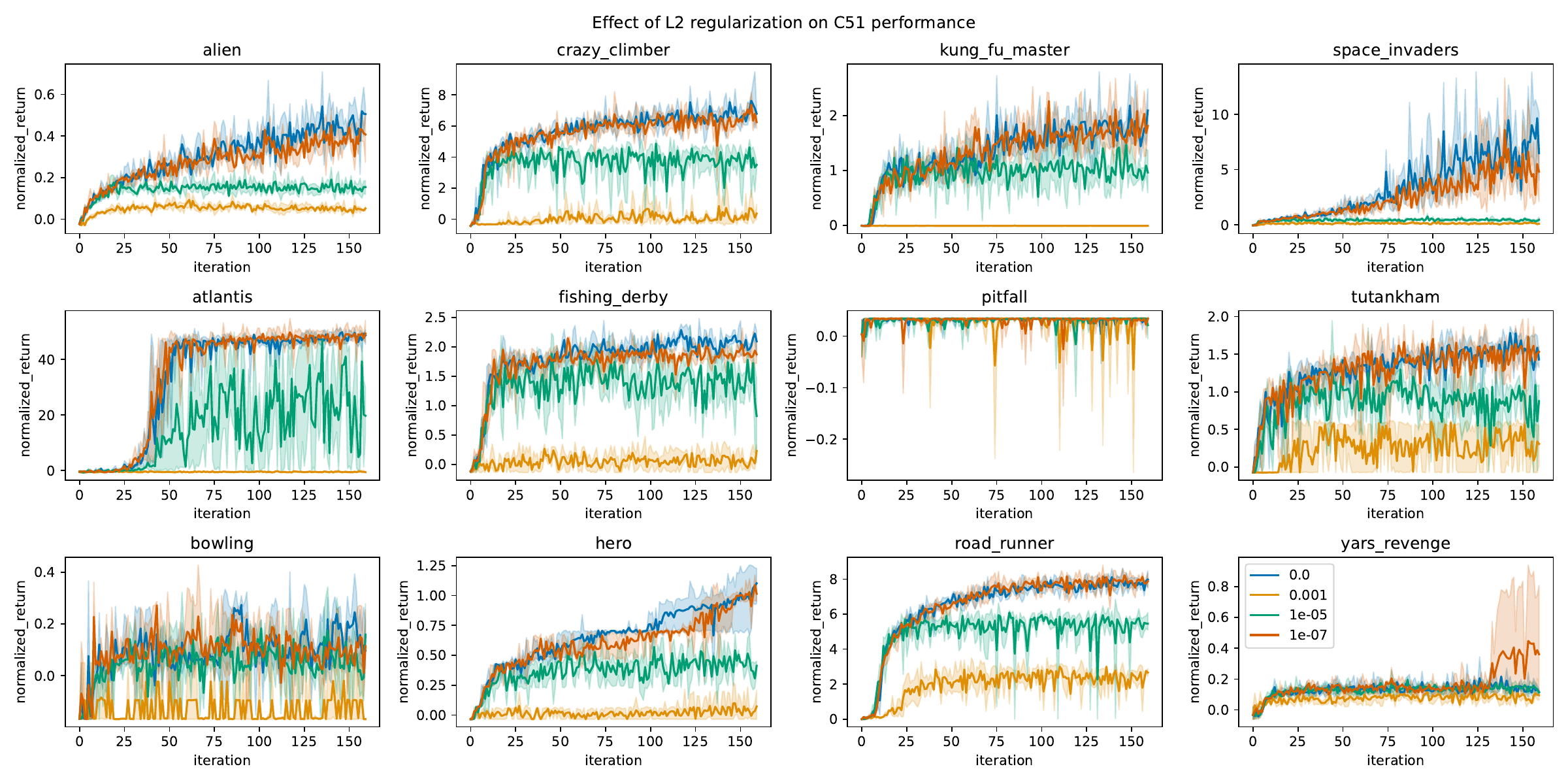}
    \caption{Parameter growth in Seaquest and effect of L2 regularization on the C51 agent in a subset of Atari. }
    \label{fig:l2-atari}
\end{figure}

We provide per-game performance curves for the deep RL agents trained on the Atari benchmark in Figure~\ref{fig:c51-per-game}, and also illustrate the nuances of applying weight regularization in this domain in Figure~\ref{fig:l2-atari}. As can be seen from these figures, layer normalization provides obvious benefits in the C51 agent, however the agent does not benefit from L2 regularization. Note that unlike the figure in the main body of the paper, our sweep over L2 regularization coefficients is only evaluated on a subset of the atari benchmark obtained by sorting the games in alphabetical order and selecting every fifth one. Our analysis of the seaquest environment shows that the parameter norm of these agents grows at a modest rate and is several orders of magnitude below the level which caused optimization challenges in the image classification tasks. We see that a penalty of 1e-7 slightly slows the rate of parameter growth without interfering with performance, a value that is much lower than what we studied in the sequential supervised learning setting. This is consistent with prior observations that deep RL agents tend not to benefit from L2 regularization.

\subsection{L2 regularization sweep}
\begin{figure}
    \centering
    \includegraphics[width=\linewidth]{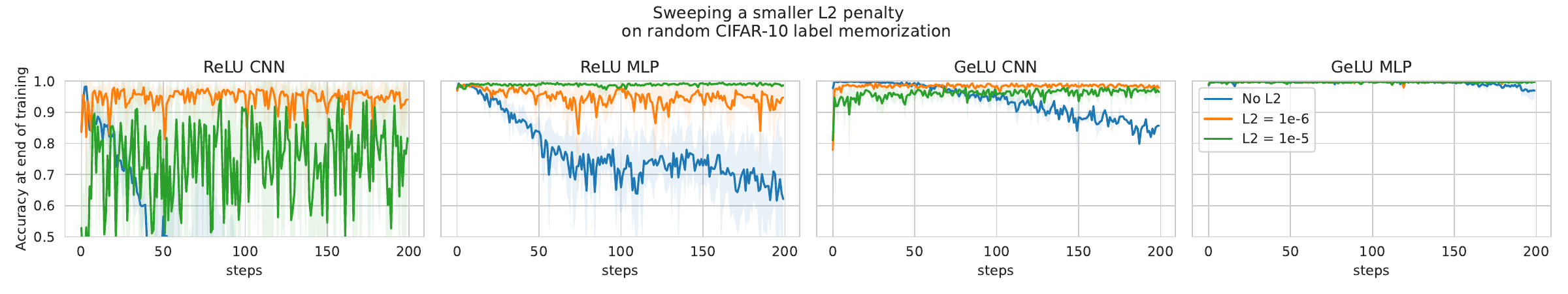}
    \caption{Comparison with a smaller L2 penalty: for suitable regularizer weights, we see minimal reduction in performance at the end of an iteration; using a larger-than-optimal regularizer can slow down training in some instances.}
    \label{fig:l2_extra_sweep}
\end{figure}

Noting a small gap in peak performance between the models trained with layer normalization but either with or without L2 regularization, we investigate whether this gap might be caused by an overly-aggressive L2 penalty. Given the highly noisy nature of the learning targets in the case of random label memorization, we would expect that maintaining small weight norm would be in tension with reducing loss on the training set, and so may require a smaller penalty than the value of $10^{-5}$ that we used as an initial `reasonable guess'. We run this experiment on four architectures (we omit the ResNet due to the longer training time required and the short rebuttal period): a depth-4 CNN with fully-connected layer width 256 and either GeLU or ReLU activations, along with a depth-4 MLP with width 512, again using either GeLU or ReLU. We run each network configuration for 200 iterations of 200K steps, and use five different random seeds. Shaded region indicates standard deviation. As we see in Figure~\ref{fig:l2_extra_sweep}, reducing the L2 penalty by an order of magnitude and allowing for a slightly longer reset interval (200,000 steps) significantly reduces the gap between regularized and unregularized convolutional networks, while for MLPs we see that the 1e-5 penalty is highly effective at maintaining plasticity while maintaining close to 100\% final performance even after 200 resets (equating to 40M optimizer steps).

\subsection{Longer run of contextual bandit task}
\begin{figure}
    \centering
    \includegraphics[width=\linewidth]{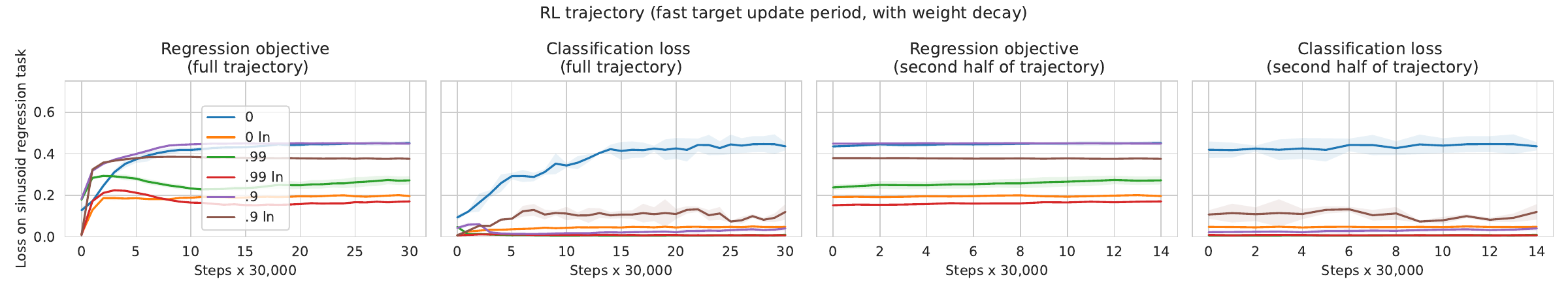}
    \includegraphics[width=\linewidth]{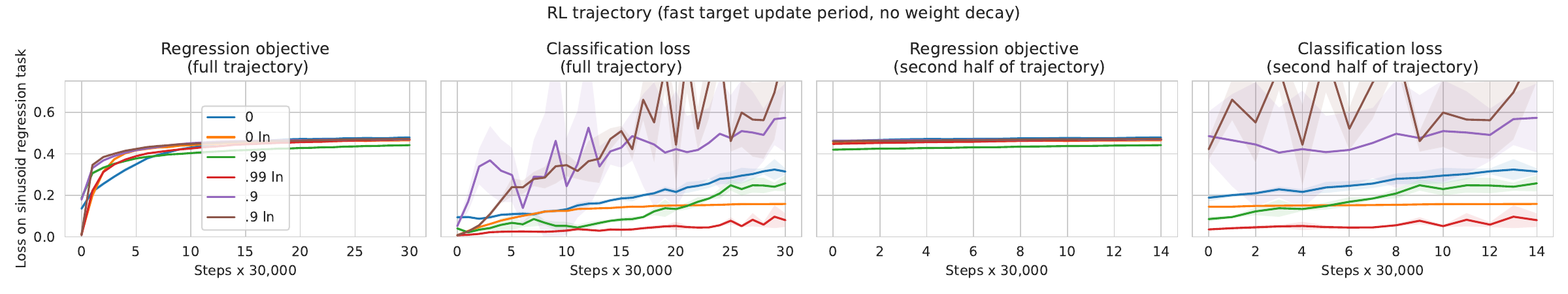}
    \includegraphics[width=\linewidth]{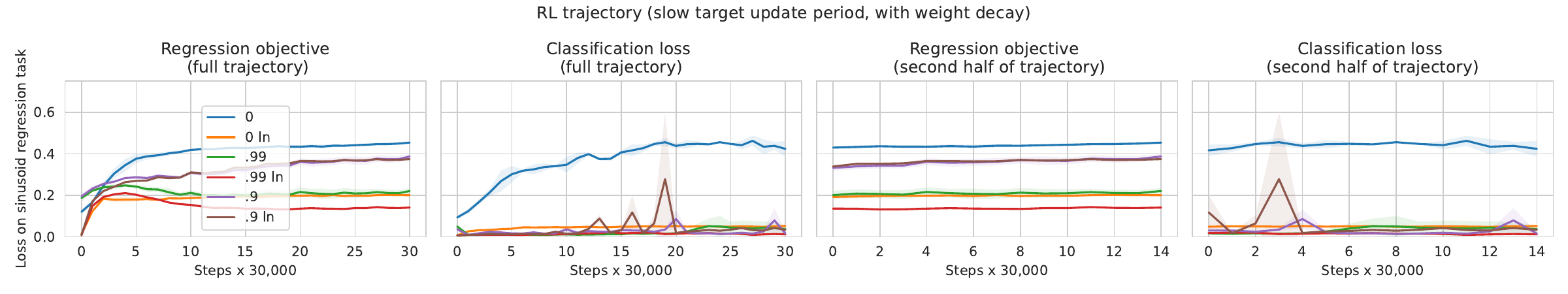}
    \includegraphics[width=\linewidth]{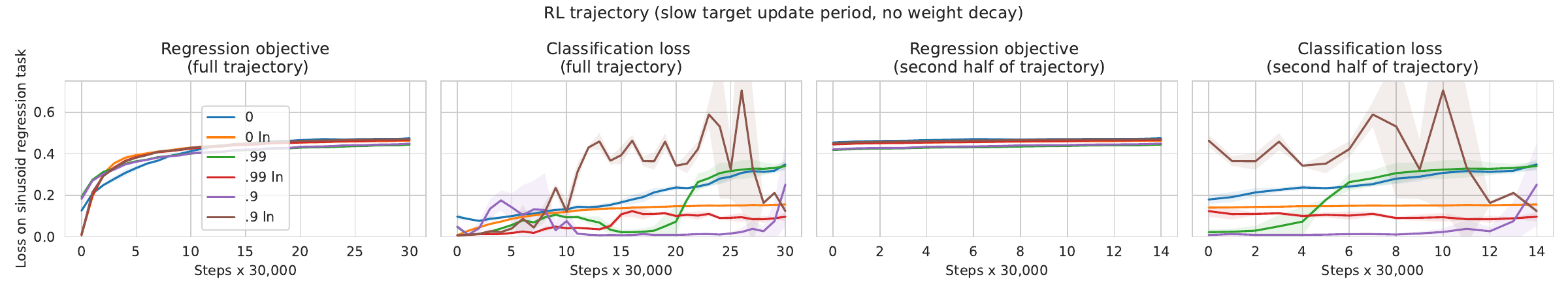}
    \caption{Evolution of plasticity under longer runs on the image classification MDP.}
    \label{fig:deep_contextual_bandit}
\end{figure}

Upon request from reviewers, we have included a longer run of the DQN agents trained on the image classification MDP in Figure~\ref{fig:deep_contextual_bandit}. The agent is trained for 500,000 steps (10x the number of iterations shown in the main body of the paper) in total using the Deep Q-learning algorithm. We sweep over configurations of the following properties: L2 regularization (1e-6 or 0), regression or classification loss, layer normalization or no layer normalization, and either a 'fast' target network update period (500 optimizer steps) or a `slow' period (5000 optimizer steps). Overall, we see that the networks trained without layer normalization, weight decay, or classification losses quickly saturate at the level of random guessing on the probe task. However, combining these approaches results in significantly better long-term performance on the probe tasks, in some cases resulting in positive forward transfer. There is some variation across configurations whereby networks may see improved or reduced performance on the probe task compared to its value at initialization, but this value tends to stabilize quickly in networks trained in the most stable regimes. In all experiments which use the classification loss, we include a label smoothing parameter of 0.1.

\subsection{WILDS dataset baseline}

We also include an additional evaluation of the iwildcam experiment including as a baseline a network which is re-initialized at each task change. Including this baseline, we see that in fact the networks trained with layer normalization not only do not exhibit plasticity loss on this task distribution, but also exhibit positive forward transfer and are able to outperform the reinitialized baseline.

\begin{figure}
    \centering
    \includegraphics[width=\linewidth]{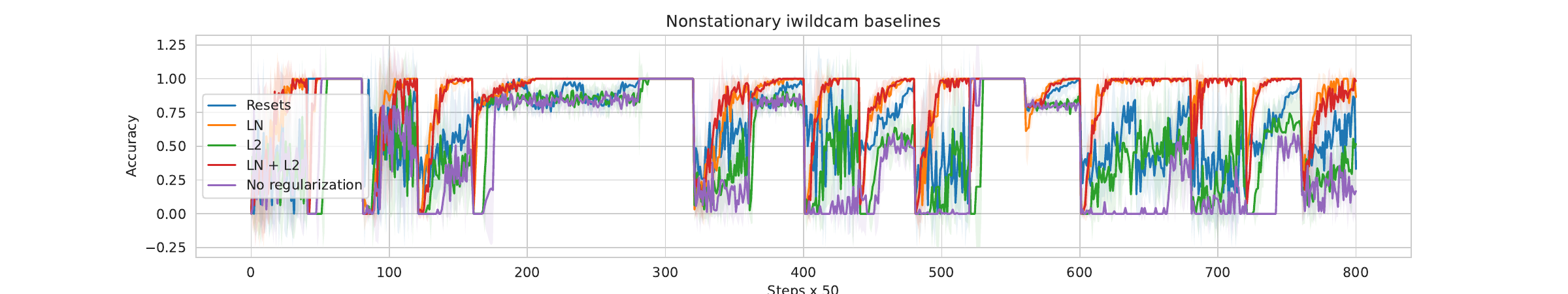}
    \includegraphics[width=\linewidth]{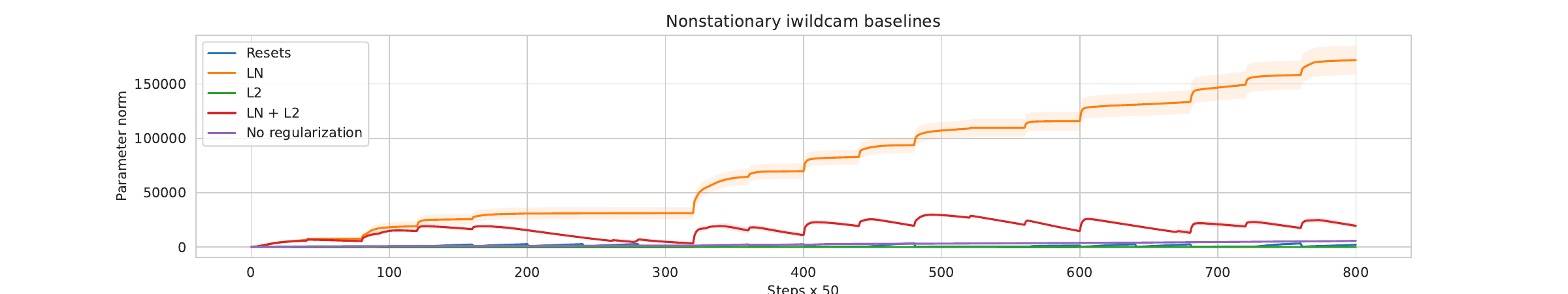}
    \caption{We see that using layer normalization allows networks to maintain plasticity and exhibit positive forward transfer on the natural distribution shift problem. The network trained with layer normalization but without L2 regularization exhibits greater parameter norm growth than the other networks, but the training run is sufficiently short that this does not manifest in training difficulties.}
    \label{fig:iwildcam-baseline}
\end{figure}

\subsection{Unit linearization in residual networks}
\label{sec:zombie-resnet}
\begin{figure}
    \centering
    \includegraphics[width=\linewidth]{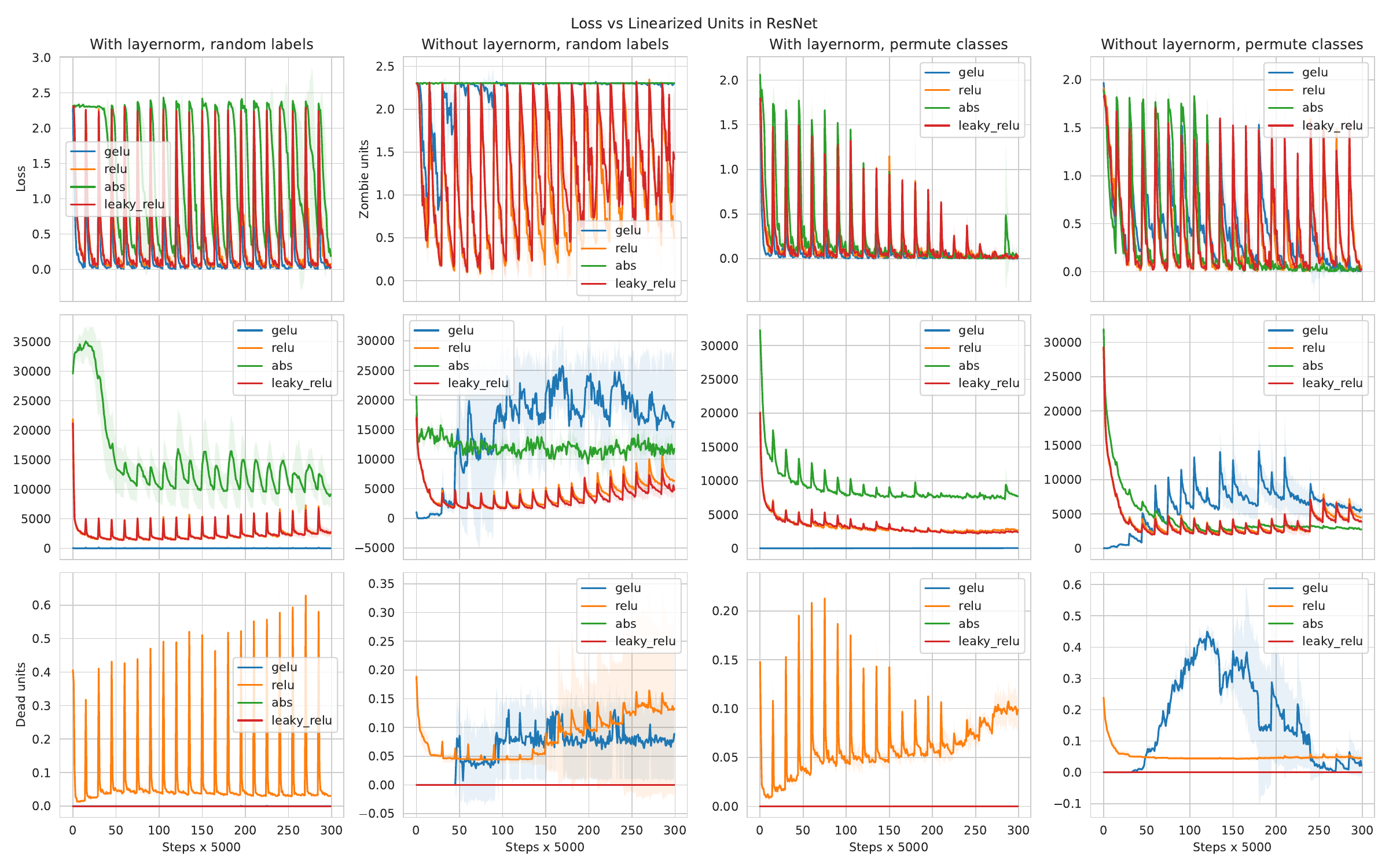}
    \caption{Loss of plasticity in our ResNet architecture trained on sequential image classification tasks. We see that high numbers of zombie units tend to correlate with networks that are unable to reduce their loss, even in networks with non-saturating activation functions such as leaky ReLU and absolute value activations. In contrast, dormant or dead neurons are only observable in the ReLU and GeLU networks, providing an incomplete picture of signal propagation failures in the network.}
    \label{fig:zombies_resnet}
\end{figure}

While we provide a hint that linearized or ``zombie'' units may be more informative than dormant or ``dead'' units in neural networks in Figure~\ref{fig:dead-and-zombies}, in fact the most compelling evidence that this phenomenon may be tied to plasticity can be observed in residual networks. The ResNet18 architecture we use is significantly deeper than the CNN and MLP, and although it features skip connections which in theory should serve as a buffer against poor signal propagation through blocks, we observe that it tends to be more sensitive to architectural factors such as layer norm than the shallower architectures. This suggests that it is also more sensitive to signal propagation failures, a feature that becomes apparent when we consider a wider range of activation functions. In Figure~\ref{fig:zombies_resnet}, we consider the ResNet18 architecture under a variety of experimental conditions on the sequential image classification regime. In particular, we consider four activation functions, two configurations of LayerNorm, and two types of nonstationarity.

We see an intriguing correlation between trends in the number of linearized units in the network and the ability of the network to reduce its loss on later tasks: networks with a large number of effectively linear units in some cases are never able to outperform random guessing, even if the network uses a non-saturating activation function and so does not suffer from dormant neurons. Further, within a trajectory, increases in the number of zombie units line up with periods where the network loss also increases, while periods where quantity declines correspond with phases where the network improves its loss (a particularly striking example of this is the case of the absolute value function with random labels and layer normalization). While raw values of the number of linearized units are not necessarily useful for cross-architecture comparisons, it is highly informative within a trajectory, and exhibits significantly stronger correlations with performance even than the number of dead units in the penultimate feature layer.

\begin{figure}
    \centering
    \includegraphics[width=0.4\linewidth]{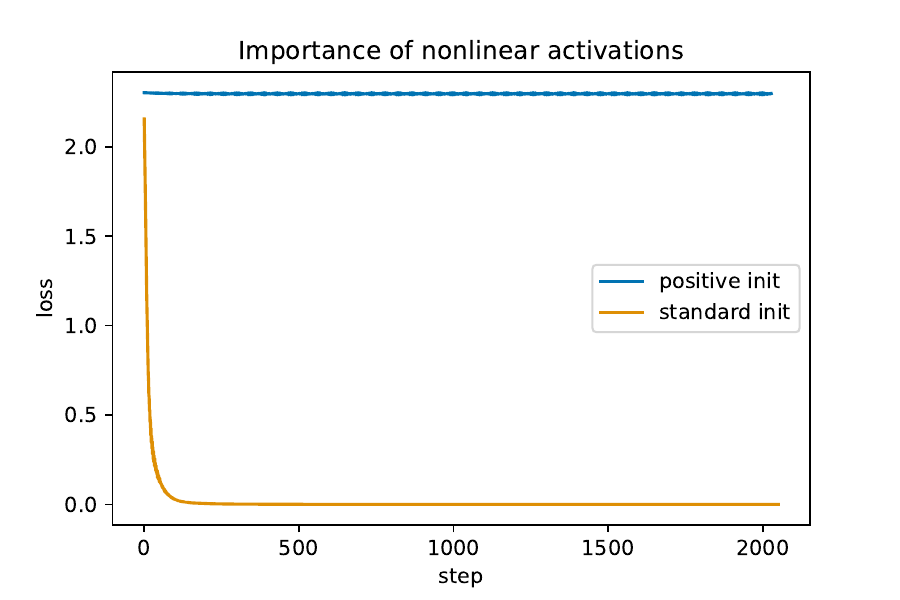}
    \caption{Effect of initializing a network to contain only zombie units in its second layer}
    \label{fig:positive_init}
\end{figure}

We also conduct a toy experiment in a small MLP trained on a subset of MNIST with canonical labels in Figure~\ref{fig:positive_init}. We first randomly initialize the network and train on the subset. We then take the same initialization and map every parameter to its absolute value. As a result, every unit in the second layer of the network is a `zombie' unit: it is still able to propagate gradients, but only applies a linear transformation to its inputs. We find that starting from this initialization, the performance of the network never decreases even after two thousand optimizer steps.

\subsection{Normalization method ablations}
\begin{figure}
    \centering
    \includegraphics[width=\linewidth]{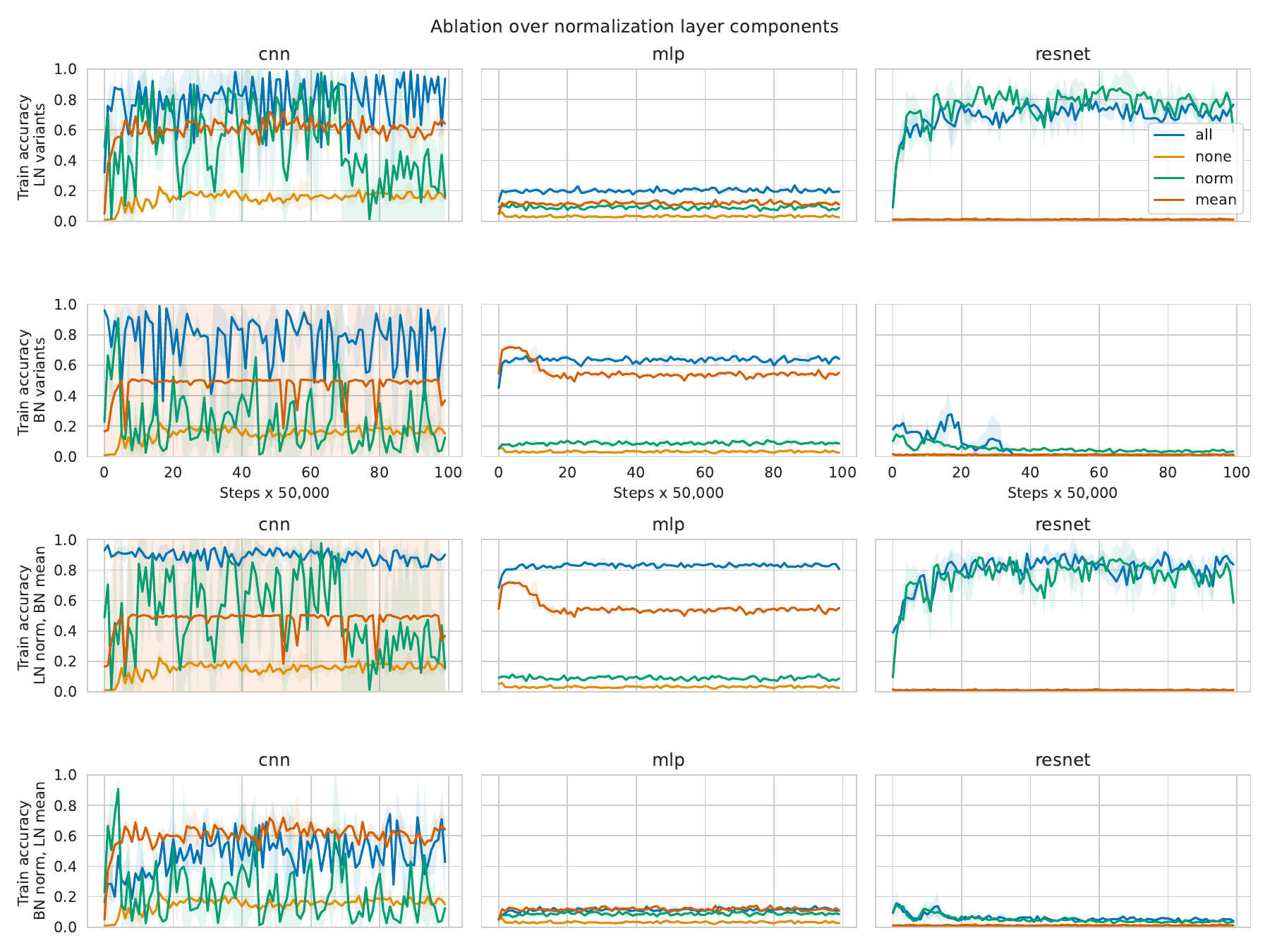}
    \caption{Ablations on the axes on which we apply different normalization transformations. We see that applying only the mean subtraction component of batch normalization along with only the standard deviation division of layer normalization results in greater robustness across architectures than applying standard layer or batch normalization. }
    \label{fig:norm-ablations}
\end{figure}
Based on the framework outlined in Figure 1, we would expect layer and batch normalization to operate on subtly different mechanisms, and therefore potentially to provide even greater benefits when used in combination. In particular, we would expect that batch normalization would provide particular benefits due to avoiding saturated units and centering the distribution of preactivations to every nonlinearity about zero, which for most activation functions is exactly where the function is most nonlinear, thus avoiding the zombie unit phenomenon. This mechanism is independent of the `normalization' component of batch norm (i.e. the division by the standard deviation of the preactivations). In contrast, while layer normalization will avoid pathologies where \textit{all} units saturate, it will have a much weaker effect on this mechanism. We would expect that the benefits of layer normalization would primarily stem from its hard constraint on the feature norm, which will have beneficial effects on signal propagation \citep{martens2021rapid} along with a normalizing effect on the feature gradients \citep{xu2019understanding}. One might then expect that a more effective strategy for maintaining plasticity, at least in supervised image classification tasks which are known to be robust to the stochasticity induced by batch normalization, would be to apply different normalization transformations along different axes of the features. 

We explore some variants of this idea in Figure~\ref{fig:norm-ablations}, where we follow the nonstationary image classification protocol described previously, using the random label memorization variant. We train each network for 50,000 steps between re-randomizations and use an L2 penalty of 1e-5. We train for 100 iterations. We see architecture-dependent effects, where for example the ResNet is not able to train in the absence of the feature norm constraint from layer normalization while the MLP struggles in the absence of mean subtraction from batch normalization. 

In the case of the MLP, we observed in later experiments that centering the inputs at approximately zero was important for training the network, and that most of the benefits of the mean subtraction from batch normalization seen here were replicated by input centering. However, we do consistently observe that across architecture, the combination of division by standard deviation along the feature axis and mean subtraction along the batch axis slightly outperforms naive layer and batch normalization.
Ultimately, because the gains we saw from this nonstandard combination were relatively minor in most architectures, and because batch normalization is known to provide much less benefit in RL or natural language tasks than it does in image classification, we opted to use standard layer normalization in most of our experiments.

\subsection{Per-game C51 and Rainbow Agent results}
\begin{figure}
    \centering
    \includegraphics[width=\linewidth]{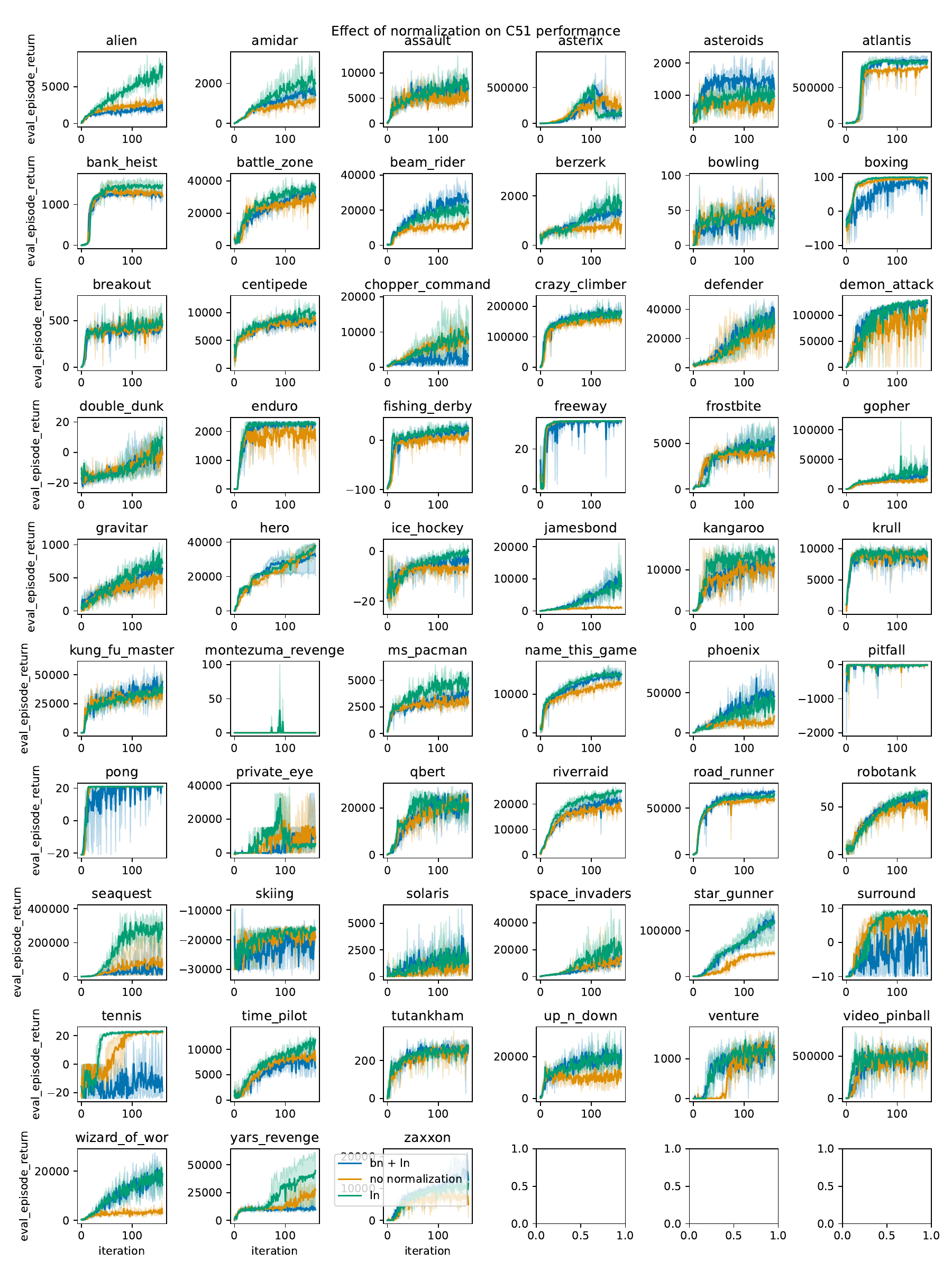}
    \caption{Per-game results for C51 agent with layer normalization or a combination of batch and layer normalization.}
    \label{fig:c51-per-game}
\end{figure}

\begin{figure}
    \centering
    \includegraphics[width=\linewidth]{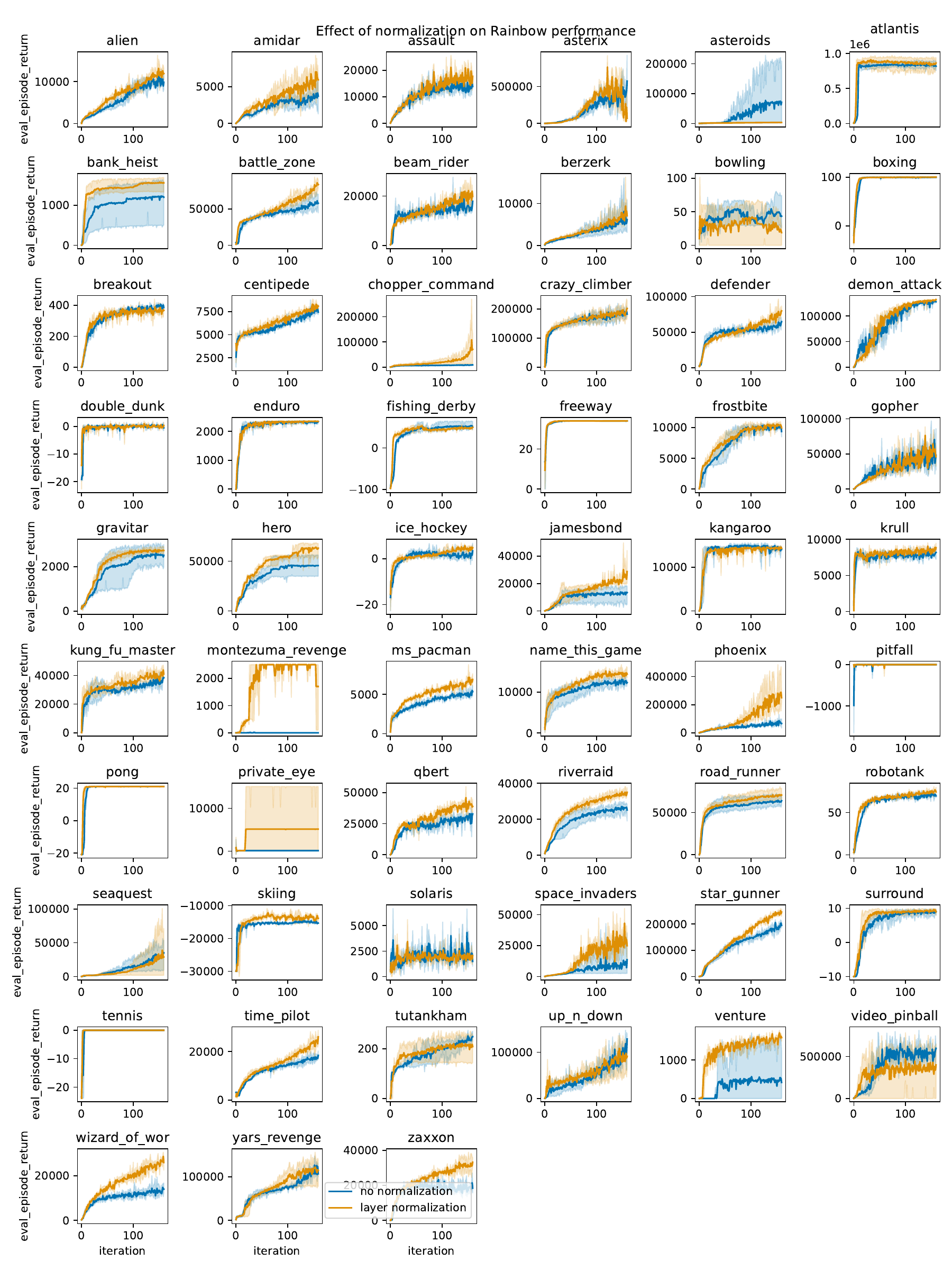}
    \caption{Per-game results for C51 agent with layer normalization or a combination of batch and layer normalization.}
    \label{fig:rainbow-per-game}
\end{figure}

We include a per-game visualization of the C51 agent's performance with and without layer normalization. We include a second variant not included in the main paper which used both batch and layer normalization, noting that our findings corroborate prior observations that batch normalization can hinder performance in deep RL as seen in Figure~\ref{fig:c51-per-game}. We also look at the Rainbow agent with and without layer normalization in Figure~\ref{fig:rainbow-per-game}.

\subsection{DeepMind Control Suite Experiment Details}
\label{sec:dm-control}
To complement our findings on Atari, we also consider a SAC agent trained on continuous control tasks from the DeepMind Control suite. This benchmark significantly differs from Atari along a number of axes: the action space is continuous, the input space is low-dimensional, and the network architectures typically used in this domain are fully-connected rather than convolutional. Somewhat surprisingly, we nonetheless see significant benefits from the incorporation of layer normalization into the architecture we used. We did not perform extensive hyperparameter tuning on this domain.

\textbf{Algorithm:} we use the Soft Actor Critic algorithm \citep{haarnoja2018soft}.

\textbf{Network:} for the critic architecture, we use a three hidden layer MLP encoder with hidden dimension 256, and put LayerNorm prior to the activations of the first- and third-layer outputs; the output of this encoder is then fed into a two-layer MLP with elu activations. The actor network architecture uses the same encoder architecture with a single 256-dimensional hidden layer prior to the Gaussian policy output head.

\textbf{Training protocol:} we train each agent for 1B environment frames and use five seeds for each experiment configuration. We use the adam optimizer with a learning rate of 3e-4 for the actor and 1e-4 for the critic.

\textbf{Domains:} we evaluate on 28 domains from the DeepMind control suite.

\textbf{Results:} we observe significant gains from layer normalization in several environments where the network constructed without normalization layers fails to make any learning progress. 

\begin{figure}
    \centering
    \includegraphics[width=\linewidth]{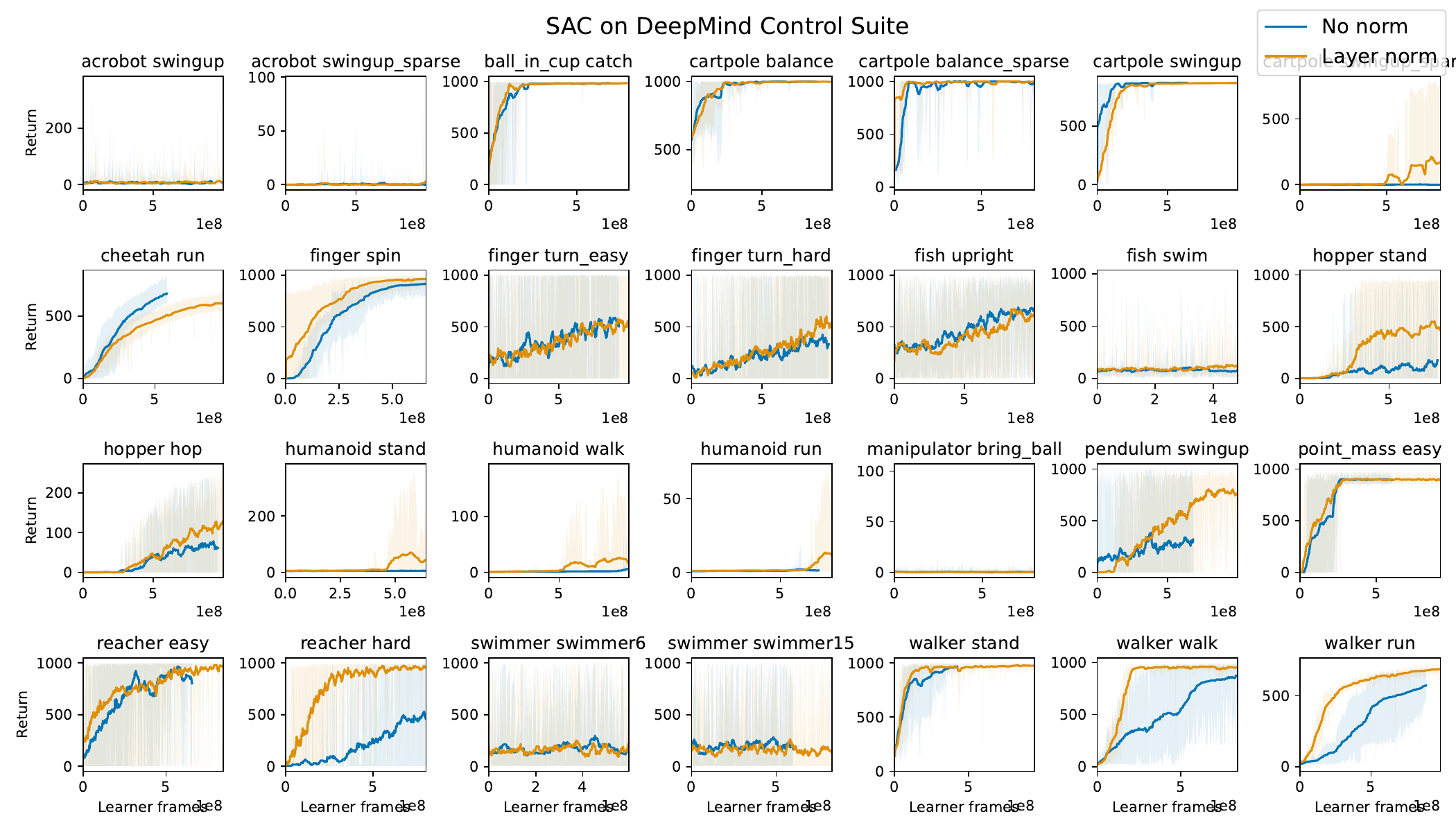}
    \caption{Effect of layer normalization in DeepMind Control Suite.}
    \label{fig:dmc-all}
\end{figure}
\section{A discussion on the independence of causal mechanisms}
\label{appx:independence}
This paper claims that the mechanisms we discuss in Section~\ref{sec:mechanisms} are \textit{independent}. To justify this claim, it is first necessary to clarify what we mean by independence. In this paper we will say that \textit{two mechanisms of plasticity loss are independent if there exist learning problems where completely mitigating one mechanism does not prevent plasticity loss due to the other.} A stronger notion of independence might require that the mechanisms not interact, however in a system as complex as training a deep neural network this is an excessively stringent requirement. 

Given this definition of independence, it is then straightforward to determine whether two mechanisms are independent, at least if the answer is positive: it suffices to identify a learning problem constructed so that one mechanism cannot occur, and where the other can be shown to cause the network to lose plasticity. We can in fact find such learning problems within the experiments presented in previous sections, which we now enumerate.

While \textbf{preactivation shift} and \textbf{weight norm} often correlate in neural network training trajectories, much of this correlation is a natural result of neural network optimization, as both weights and features must necessarily change over the course of learning, and any two quantities which consistently trend away from their initial values may look correlated. To see that the mechanisms by which these phenomena influence plasticity are independent, we first consider a network where the weight norm is regularized via weight decay and can be verified not to grow uncontrollably, but where preactivation shift causes performance degradation. We find such an example in Figure~\ref{fig:iwildcam-baseline}, where the network trained with L2 regularization exhibits declining performance, while the network trained with L2 + layer normalization does not. Similarly, the CNN shown in Figure 8 exhibits loss of plasticity in long training trajectories when only layer normalization is applied, but not when layer normalization is used in conjunction with L2 regularization.

In contrast, the relationship between preactivation shift and \textbf{unit saturation} is more of a direct causal arrow: it is possible for pathologies to arise as a result of preactivation shift that do not correspond to unit death, for example the linearized units illustrated in Figure 19; however, unit death requires a shift in the distribution of preactivations in order to occur (a counter example would be when units are dormant at initialization, however in this case the problem is not loss of plasticity but rather its nonexistence). Intriguingly, the results of Figure 21 do suggest that, all else being equal, a shift in the distribution of preactivations which preserves the per-unit mean over samples is less damaging than one which preserves the per-sample mean over units, consistent with the unsurprising conclusion that zeroing out gradient flow through a feature is more damaging than other more subtle signal propagation issues.

\textbf{Regression target magnitude} is a unidirectional factor in plasticity loss, as in most cases the magnitude of the regression targets is not caused by other optimization pathologies in the network, but rather is an unavoidable feature of the learning problem. The magnitude of regression targets can exacerbate \textbf{preactivation shift}, as large targets require commensurate changes in the network to fit. However, we see that even in networks which feature layer normalization, for which the preactivation mean and variance are fixed, we still see reduced ability to adapt to new tasks as a function of pretraining target magnitude in Figure~\ref{fig:regression-finetuning-scale}. Similarly, while regression target magnitude can exacerbate \textbf{weight norm} growth, only adding L2 regularization is insufficient to mitigate the issue, as we can see in the RL agents in Figure 3 (RHS); similarly, the incomplete success of layer normalization at improving plasticity in this domain suggests that avoiding unit saturation across an entire layer is insufficient to mitigate plasticity loss caused by large target magnitudes. 

\textbf{Weight norm growth:} it remains to show the independence of parameter norm growth and number of dead units as causes of plasticity loss. This is a relatively straightforward task: weight norm growth is often most pronounced in networks with layer normalization (see e.g. Figure 4), which avoid at least the most extreme version of unit dormancy where an entire layer becomes dormant. In the opposite direction, unit saturation is a highly effective means of reducing the growth of the parameter norm as it prevents gradients from acting on parameters in a way that could increase their magnitude, however has obvious deleterious effects on trainability.